\documentclass[10pt, journal]{IEEEtran}

\usepackage{cite}
\usepackage{amsmath,graphicx}
\usepackage{epstopdf}
\usepackage{ragged2e}
\usepackage{xcolor}
\usepackage{amssymb}
\usepackage{pifont}
\usepackage{array}
\usepackage{booktabs}
\usepackage{bm}
\usepackage{multirow}
\usepackage{graphicx}
\usepackage{color}
\usepackage{float}
\usepackage{stfloats}
\usepackage[misc]{ifsym}
\usepackage{stmaryrd}
\usepackage{grffile}
\usepackage{subfig}
\usepackage{makecell}
\usepackage{threeparttable}
\usepackage{enumitem}
\usepackage{enumerate}
\usepackage[pagebackref=false,breaklinks=true,letterpaper=true,colorlinks,bookmarks=false]{hyperref}
\usepackage{colortbl}
\definecolor{shadow}{cmyk}{.08,0,0,0}
\usepackage{tikz}
\newcommand*{\circled}[1]{\lower.6ex\hbox{\tikz\draw (0pt, 0pt)%
		circle (.4em) node {\makebox[1em][c]{\tiny #1}};}}
\begin{document}

\title{Real-World Light Field Image Super-Resolution via Degradation Modulation}

\author{Yingqian Wang, Zhengyu Liang, Longguang Wang, Jungang Yang, Wei An, Yulan Guo

\IEEEcompsocitemizethanks{
\IEEEcompsocthanksitem
Y. Wang, Z. Liang, L. Wang, J. Yang, W. An and Y. Guo are with the College of Electronic Science and Technology, National University of Defense Technology, Changsha, P. R. China.
}
\thanks{This work was partially supported by National Key R\&D Program of China (No. 2021YFB3100800) and National Natural Science Foundation of China (Nos. U20A20185, 61972435, 61921001). Corresponding author: Jungang Yang. Emails: yangjungang@nudt.edu.cn.}
}

\markboth{}
{Shell \MakeLowercase{\textit{et al.}}: Bare Advanced Demo of IEEEtran.cls for IEEE Computer Society Journals}

\IEEEtitleabstractindextext{
\begin{abstract}
\justifying
 Recent years have witnessed the great advances of deep neural networks (DNNs) in light field (LF) image super-resolution (SR). However, existing DNN-based LF image SR methods are developed on a single fixed degradation (e.g., bicubic downsampling), and thus cannot be applied to super-resolve real LF images with diverse degradation. In this paper, we propose a simple yet effective method for real-world LF image SR. In our method, a practical LF degradation model is developed to formulate the degradation process of real LF images. Then, a convolutional neural network is designed to incorporate the degradation prior into the SR process. By training on LF images using our formulated degradation, our network can learn to modulate different degradation while incorporating both spatial and angular information in LF images. Extensive experiments on both synthetically degraded and real-world LF images demonstrate the effectiveness of our method. Compared with existing state-of-the-art single and LF image SR methods, our method achieves superior SR performance under a wide range of degradation, and generalizes better to real LF images. Codes and models are available at \url{https://yingqianwang.github.io/LF-DMnet/}.
\end{abstract}

}

\maketitle

\IEEEdisplaynontitleabstractindextext

\IEEEpeerreviewmaketitle

\section{Introduction}\label{sec:introduction}

\IEEEPARstart{L}{ight} field (LF) cameras record both intensity and direction of light rays, and enable many applications such as refocusing \cite{jayaweera2020multi}, depth estimation \cite{OACC-Net, EPINET, zhou2023beyond}, and view rendering \cite{zhu2020occlusion,wu2021revisiting,SAAN}.
 Since high-resolution (HR) LF images are beneficial to various applications but are generally obtained at an expensive cost, it is necessary to reconstruct HR LF images from low-resolution (LR) LF images, i.e., to achieve LF image super-resolution (SR).

In the past decade, deep neural networks (DNNs) have been successfully applied to LF image SR and achieved significant progress \cite{zhang2022heat,tian2022heterogeneous,hu2020meta,li2022local,wu2023practical}. In the area of LF image SR, many networks \cite{LFCNN15,LFCNN17,LFNet,LFSSR,HDDRNet,cheng2019light,resLF,ATO,LF-InterNet,LF-DFnet,MEG-Net,ZSLFSR,cheng2022spatial,AFR,DistgLF,DPT,LFT,EPIT} were developed to improve SR accuracy. However, real-world LF image SR has remained under investigated due to the following two reasons. First, it is challenging to develop an LF image SR model that can handle real-world degradation. Real-world LF images suffer from diverse degradation which varies with both imaging devices (e.g., Lytro or RayTrix cameras) and shot conditions (e.g., scene depth, focal length, illuminance). However, existing LF image SR methods focus on the design of network architecture, and develop models on the simple bicubic downsampling degradation. Consequently, these methods suffer a notable performance drop when applied to real LF images. Second, it is challenging to simultaneously utilize the degradation information while incorporating the complementary angular information. Existing methods generally achieve real-world SR on single images (i.e., ignore the view-wise correlation), and thus cannot achieve satisfactory performance on LF image SR.

 \begin{figure}
 \centering
 \includegraphics[width=8.8cm]{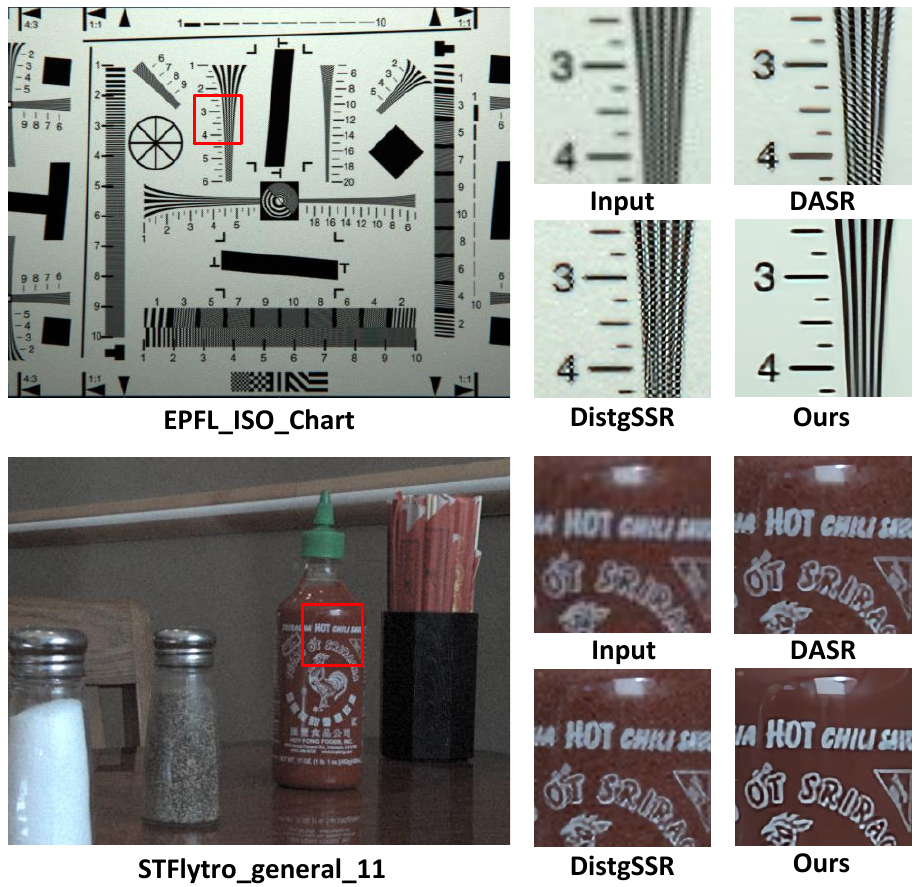}
 \caption{Visual results achieved by DASR \cite{DASR}, DistgSSR \cite{DistgLF} and our method on real LF images for $4\times$SR. Scenes \textit{ISO\_Chart} from the EPFL dataset \cite{EPFL} and \textit{general\_11} from the STFlytro dataset  \cite{STFlytro} are used for comparison.}
 \label{fig:thumbnail}
 \end{figure}

In this paper, we propose a simple yet effective method for real-world LF image SR. In our method, we first formulate a practical degradation model to approximate the degradation process of real LF images, and then develop a convolutional neural network to super-resolve LF images with diverse and real degradation. To incorporate the degradation prior into the SR process, we design a degradation-modulating convolution (DM-Conv) whose weights are dynamically generated according to the degradation representation. By integrating the proposed DM-Conv with the disentangling mechanism \cite{DistgLF}, our network (namely, LF-DMnet) can well incorporate spatial and angular information under diverse degradation. As shown in Fig.~\ref{fig:thumbnail}, compared with DistgSSR \cite{DistgLF} and DASR \cite{DASR}, our method achieves better performance on real LF images and generates images with more clear details and fewer artifacts.

 The contributions of this work are summarized as follows:

 \begin{itemize}
 \item We propose a practical LF degradation model to handle the real-world LF image SR problem. Different from existing works which focus on the advanced network designs, we firstly address the importance of degradation formulation and modulation in LF image SR.
 \item We propose a degradation-modulating network (i.e., LF-DMnet) to incorporate the degradation prior into the SR process. Extensive ablation studies and model analyses validate the effectiveness of our degradation modulation mechanism.
 \item Our method achieves state-of-the-art SR performance on both synthetic and real-world degradation, which not only provides a simple yet strong baseline, but also takes a step towards practical real-world LF image SR.
\end{itemize}

 The rest of this paper is organized as follows. In Section~\ref{sec:RelatedWork}, we briefly review the related works. In Section~\ref{sec:Degradation}, we describe our degradation model for LF image SR. In Section~\ref{sec:Network}, we introduce the details and design thoughts of our LF-DMnet. Experimental results are presented in Section~\ref{sec:Experiment}. Finally, we conclude this paper in Section \ref{sec:Conclusion}.

\section{Related Work}\label{sec:RelatedWork}
In this section, we briefly review several major works for DNN-based single image SR and LF image SR.

\subsection{Single Image Super-Resolution}
The goal of single image SR is to reconstruct an HR image from its LR version. According to different degradation settings, existing single image SR methods can be roughly categorized to single degradation based methods and multi-degradation based methods.

Early works on DNN-based single image SR are generally developed on a single and fixed degradation (e.g., bicubic downsampling). Dong et al. \cite{SRCNN14} firstly applied convolution neural networks to image SR and developed a three-layer network named SRCNN. Although SRCNN is shallow and lightweight, it outperforms many traditional SR methods \cite{timofte2013anchored,timofte2014a+,yang2010image,kim2010single}. Since then, deep networks have dominated the SR area and achieved continuously improved accuracy with large models and complex architectures. Kim et al. \cite{VDSR} applied global residual learning strategy to image SR and developed a twenty-layer network called VDSR. Lim et al. \cite{EDSR} proposed an enhanced deep SR (EDSR) network by using both global and local residual connections. Zhang et al. \cite{RDN} combined residual learning with dense connection to build a residual dense network with more than 100 layers. Subsequently, Zhang et al. \cite{RCAN} developed a very deep network in a residual-in-residual architecture to achieve competitive SR accuracy. More recently, attention mechanism \cite{RNAN,DNLN,zhang2021context} and Transformer architectures \cite{SwinIR,lu2022transformer} have been extensively studied to achieve state-of-the-art SR performance.

Although the aforementioned methods have achieved continuously improved SR performance, they are designed for a single fixed degradation (e.g., bicubic downsampling) and will suffer from a significant performance drop when the degradation differs from the assumed one. Consequently, many methods have been proposed to achieve image SR with multiple various degradation \cite{blindsr-survey}. Zhang et al. \cite{SRMD} proposed an SRMD network where the degradation map was concatenated with the LR image as the input of the DNN. Subsequently, Xu et al. \cite{UDVD} applied dynamic convolutions to achieve better SR performance than SRMD. In \cite{USRnet}, an unfolding SR network was developed to handle different degradation by alternately solving a data sub-problem and a prior sub-problem. Gu et al. \cite{IKC} proposed an iterative kernel correction method (namely, IKC) to correct the estimated degradation by observing previous SR results. More recently, Wang et al. \cite{DASR} achieved degradation representation learning in a contrastive manner and developed a degradation-aware SR network named DASR for real-world single image SR.

\subsection{Light Field Image Super-Resolution}
The goal of LF image SR is to super-resolve each sub-aperture image (SAI) of an LF. A straight-forward scheme to achieve LF image SR is applying single image SR methods to each SAI independently. However, this scheme cannot achieve a good performance since the complementary angular information among different views is not considered. Consequently, existing LF image SR methods focus on designing advanced network architectures to fully use both spatial and angular information.

\begin{figure*}[t]
	\centering
	\includegraphics[width=16cm]{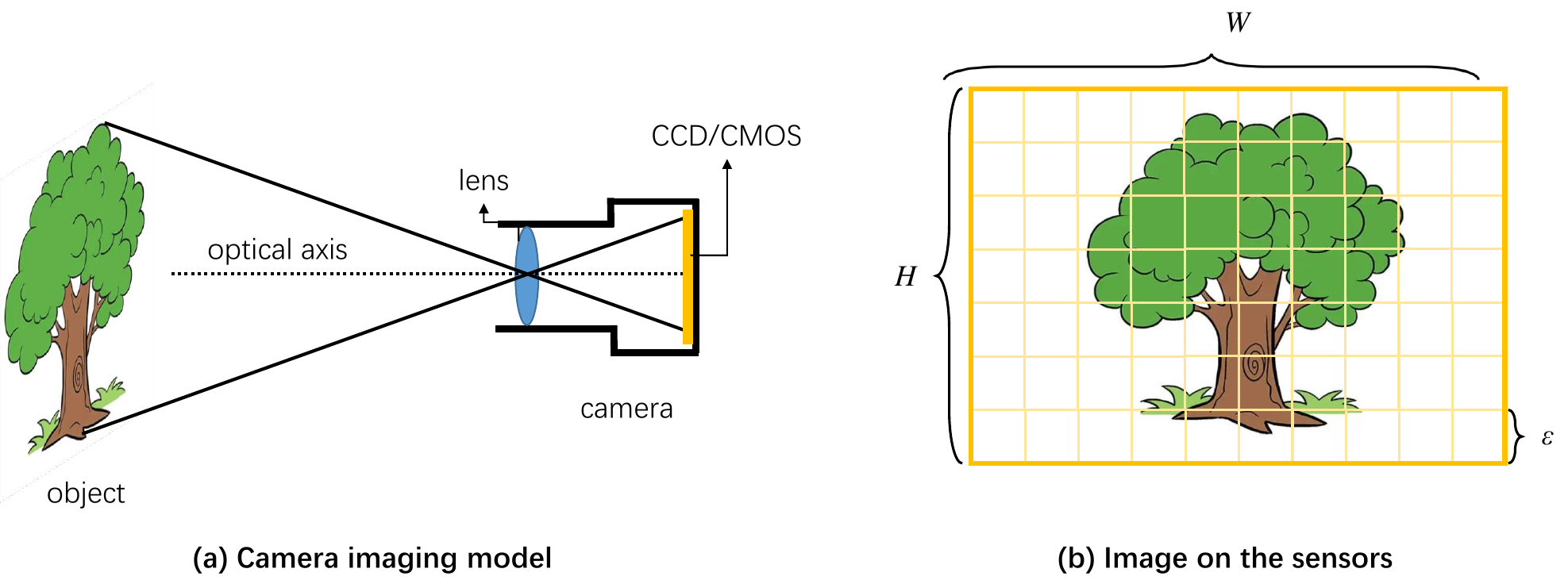}
	\vspace{0.1cm}
	\caption{An illustration of the camera imaging process.}\label{fig:camera-model}
\end{figure*}

Yoon et al. \cite{LFCNN15} proposed the first DNN-based method called LFCNN to enhance both spatial and angular resolution of an LF. In their method, SAIs are first super-resolved using SRCNN \cite{SRCNN14}, and then finetuned in pairs or quads to incorporate angular information. Wang et al. \cite{LFNet} proposed a bidirectional recurrent network for LF image SR, in which the angular information in adjacent horizontal and vertical views was incorporated in a recurrent manner. Zhang et al. \cite{resLF} proposed a multi-branch residual network to incorporate the multi-directional epipolar geometry prior for LF image SR. In their subsequent work MEG-Net \cite{MEG-Net}, the SR performance was further improved by applying 3D convolutions to SAI stacks of different angular directions. Jin et al. \cite{ATO} developed an all-to-one method for LF image SR, and performed structural consistency regularization to preserve the LF parallax structure. Wang et al. \cite{LF-InterNet} developed an LF-InterNet to repetitively interact spatial and angular information for LF image SR, and then generalized the spatial-angular interaction mechanism to the disentangling mechanism \cite{DistgLF} to achieve state-of-the-art SR accuracy.

More recently, Wang et al. \cite{LF-DFnet} used deformable convolutions \cite{deformable,deformableV2} to address the disparity problem in LF image SR. Cheng et al. \cite{ZSLFSR} proposed a zero-shot learning scheme to handle the domain gap among different LF datasets. Liang et al. \cite{LFT} proposed a Transformer-based LF image SR network, in which a spatial Transformer and an angular Transformer were designed to model long range spatial dependencies and angular correlation, respectively. Wang et al. \cite{DPT} proposed a detail-preserving Transformer to exploit non-local context information and preserve details for LF image SR. Liang et al. \cite{EPIT} investigated the non-local spatial-angular correlations in LF image SR, and developed a Transformer-based network called EPIT to achieve state-of-the-art SR performance.


Although remarkable progress have been achieved in LF image SR, existing methods only focus on the advanced network design but ignored the generalization capability to real-world degradation. In this paper, we handle the real-world LF image SR problem by formulating a practical LF degradation model and designing a degradation-modulating network.

\section{LF Image Degradation Formulation}\label{sec:Degradation}
 In this section, we formulate a general and practical degradation model for real-world LF image SR. In Sec.~\ref{sec:DegradationFormulation}, we analyze the camera imaging process and derive the image degradation model. In Sec.~\ref{sec:LFDegradationModel}, we extend the degradation model to 4D LF images to build the LF image degradation model, and discuss its key components. In Sec.~\ref{sec:DegradationDiscuss}, we compare the differences between our method and existing SR methods.

 \subsection{Degradation Formulation}\label{sec:DegradationFormulation}

 In this section, we first formulate the camera imaging process considering three key factors including \textit{point spread function (PSF)}, \textit{sensor sampling} and \textit{additional noise}. Then, we derive the image degradation model based on the formulated camera imaging process.

 Figure~\ref{fig:camera-model} shows a toy example of the camera imaging process, in which the light rays are firstly projected onto the sensor plane (as shown in Fig.~\ref{fig:camera-model}(a)), and then sampled by the sensor units (as shown in Fig.~\ref{fig:camera-model}(b)).
 Let $\mathcal{I}_\textit{real}:(x,y) \rightarrow \mathbb{R}$ be the real image (a 2D continuous function) on the sensor plane, $k_\textit{psf}$ be the point spread function of the camera imaging system\footnote{According to the signal processing theory, PSF can be considered as the \textit{unit impulse response} of the camera imaging system.}, $\mathcal{I}_\textit{ideal}:(x,y) \rightarrow \mathbb{R}$ be the ``ideal'' image (a 2D continuous function) without considering the point spread process. According to the camera imaging process, the ``real'' image is obtained by convolving the ``ideal'' image with the point spread function, i.e., 
 \begin{equation}\label{eq:1}
 	\mathcal{I}_\textit{real}(x, y) = \int_{-\infty}^{+\infty} \int_{-\infty}^{+\infty} k_\textit{psf}(u,v) \cdot \mathcal{I}_\textit{ideal}(x-u, y-v) ~dudv,
 \end{equation}
 which can be denoted as
 \begin{equation}\label{eq:2}
 	\mathcal{I}_\textit{real} = \mathcal{I}_\textit{ideal} \otimes k_\textit{psf},
 \end{equation}
 where $\otimes$ represents the convolution operation. Assume that the size of each sensor unit is $\epsilon \times \epsilon$, the sampling process on the sensor unit $(h,w)$ can be formulated as
 \begin{equation}\label{eq:3}
 	\mathcal{I}_\textit{LR}(h, w) = \int_{h-\frac{\epsilon}{2}}^{h+\frac{\epsilon}{2}} \int_{w-\frac{\epsilon}{2}}^{w+\frac{\epsilon}{2}} \mathcal{I}_\textit{real}(x, y) dxdy + \mathcal{N}(h, w),
 \end{equation}
 where $mathcal{I}_\textit{LR}\in \mathbb{R}^{H\times W}$ is the output of the sensor (i.e., a digital image). Here, we introduce $\left[ \cdot \right]_{\epsilon}$ to denote the sampling process with a sampling grid of $\epsilon \times \epsilon$. Then, Eq.~\ref{eq:3} can be rewritten as 
 \begin{equation}\label{eq:4}
 	\mathcal{I}_\textit{LR} = \left[ \mathcal{I}_\textit{real} \right]_{\epsilon} + \mathcal{N},
 \end{equation}
 where $\mathcal{N}\in \mathbb{R}^{H\times W}$ represents random noise in the imaging process.
 
 In image SR task, it is expected to reconstruct (or estimate) the ideal image function $\mathcal{I}_\textit{ideal}$ from the observed LR image $mathcal{I}_\textit{LR}$. Since continuous 2D image function needs to be presented via a digital image, we further introduce HR image $\mathcal{I}_\textit{HR}\in \mathbb{R}^{\alpha H \times \alpha W}$ to quantize $\mathcal{I}_\textit{ideal}$, i.e., 
 \begin{equation}\label{eq:5}
 	\mathcal{I}_\textit{HR} = \left[ \mathcal{I}_\textit{ideal} \right]_{\frac{\epsilon}{\alpha}},
 \end{equation}
 where $\alpha$ is defined as the upsampling factor. From Eq.~\ref{eq:5}, we can consider that the HR image is obtained by sampling the ideal image $\mathcal{I}_\textit{ideal}$ with smaller interval $\frac{\epsilon}{\alpha} \times \frac{\epsilon}{\alpha}$. Here, we introduce a downsampling operator $\left(\cdot \right)_{\downarrow_{\alpha}}$ to build the relationship between LR image and its HR version, i.e., 
 \begin{equation}\label{eq:6}
 	\left[ \mathcal{I} \right]_{\epsilon} = \left( \left[ \mathcal{I} \right]_{\frac{\epsilon}{\alpha}}\right)_{\downarrow_{\alpha}}, ~~~~ \mathcal{I}=\mathcal{I}_\textit{ideal} ~or~ \mathcal{I}_\textit{real}
 \end{equation}
 Substitute Eq.~\ref{eq:2} and Eq.~\ref{eq:6} into Eq.~\ref{eq:4}, we can obtain
 \begin{equation}\label{eq:7}
 	\mathcal{I}_\textit{LR} = \left( \left[ \mathcal{I}_\textit{ideal} \otimes k_\textit{psf} \right]_{\frac{\epsilon}{\alpha}}\right)_{\downarrow_{\alpha}} +  \mathcal{N}.
 \end{equation}
 According to the \textit{commutative law of convolution and sampling} (see Appendix\footnote{\url{https://yingqianwang.github.io/LF-DMnet/Appendix.pdf}} for prove), there is
 \begin{equation}\label{eq:8}
 	\left[ \mathcal{I}_\textit{ideal} \otimes k_\textit{psf} \right]_{\frac{\epsilon}{\alpha}}
 	=  \left[ \mathcal{I}_\textit{ideal} \right]_{\frac{\epsilon}{\alpha}} \otimes k_\textit{psf}.
 \end{equation}
 When we substitute Eq.~\ref{eq:8} into Eq.~\ref{eq:7}, we can obtain the image degradation model as
 \begin{equation}\label{eq:9}
 	\mathcal{I}_\textit{LR} =  \left( \mathcal{I}_\textit{HR} \otimes k \right)_{\downarrow_{\alpha}} + \mathcal{N}.
 \end{equation}
 The above degradation model can be considered as a process in which the real observed LR image is obtained by blurring, downsampling and adding noise on the HR image. In the following subsection, we will apply the degradation model to 4D LFs to formulate our LF image degradation model.

\subsection{LF Image Degradation Model}\label{sec:LFDegradationModel}
 
  We use the two-plane model \cite{levoy1996light} to parameterize 4D LF as $\mathcal{L} \in \mathbb{R}^{U\times V \times H\times W}$, where $U$ and $V$ represent angular dimensions, $H$ and $W$ represent spatial dimensions. Since this paper focuses on enhancing the spatial resolution of LFs, we use the SAI representation in \cite{DistgLF} to describe our method. That is, an LF can be considered as a $U\times V$ array of SAIs, and each SAI has a spatial size of $H\times W$.

 Here, we extend our degradation model (i.e., Eq.~\ref{eq:9}) to 4D LFs and build the LF image degradation model as
 \begin{equation}\label{eq:degrad}
     \mathcal{I}^{\textit{lr}}_{u,v} = (\mathcal{I}^{\textit{hr}}_{u,v} \otimes {k}_{u,v}){\downarrow}_{\alpha} + \mathcal{N}_{u,v},
 \end{equation}
 where $\mathcal{I}^{\textit{lr}}_{u,v} \in \mathbb{R}^{H\times W\times3}$ denotes the input LR SAI of view $(u, v)$, and $\mathcal{I}^{\textit{hr}}_{u,v} \in \mathbb{R}^{\alpha H\times \alpha W\times3}$ denotes the corresponding HR SAI. $k_{u,v} \in \mathbb{R}^{21 \times 21}$ and $\mathcal{N}_{u,v} \in \mathbb{R}^{H\times W\times3}$ represent the blur kernel and additional noise of view $(u,v)$, respectively. In the following text, we introduce the details of the three key components (i.e., blur kernel, noise, and downsamping) of our LF image degradation model.

 \subsubsection{Blur Kernel}
 We follow existing works \cite{SRMD,IKC} to use the isotropic Gaussian kernel parameterized by kernel width to synthesize blurring LF images. Note that, although anisotropic kernels (e.g., anisotropic Gaussian blur and motion blur) are also used in recent single image SR methods \cite{DASR,FKP,MADNet,DPSR,USRnet} for degradation modeling, we do not consider these blur kernels in our method because under LF structures, the rotation angle of the anisotropic Gaussian kernel and the trajetory of the motion blur of each SAI should be different but correlated. The formulation of these anisotropic blur kernels depends on the 6D pose changing of LF cameras, and belongs to the LF deblurring task \cite{srinivasan2017light,lee2018joint,mohan2018divide}. As demonstrated in Section~\ref{sec:comparison-real}, based on the isotropic Gaussian blur assumption, our method can achieve promising SR performance on real LF images.

 \subsubsection{Noise}
 Real-world LF images (especially those captured by Lytro cameras) generally have large noise. Directly super-resolving noisy LF images without performing noise reduction can result in visually unpleasant artifacts (see Section~\ref{sec:mismatch-real}). In this paper, we consider the simple channel-independent additive white Gaussian noise in our degradation process. Each element in the noise tensor $\mathcal{N} \in \mathbb{R}^{H \times W \times 3}$ is a random variable with a mean value of $0$ and an adjustable standard deviation (i.e., noise level). It is demonstrated in Section~\ref{sec:mismatch-real} that, when the noise term is considered in our degradation model, the super-resolved images are more smooth and clean with less noise residual and ringing artifacts.

 \subsubsection{Downsampling}
 We adopt the widely-used bicubic downsampling approach in our method. In this way, our degradation model can be degeneralized to a standard bicubic downsampling degradation when the kernel width and noise level equal to zero. Note that, different from blur kernel and noise level which can vary in the training phase, the downsampling approach is assumed to be fixed.

 \begin{figure*}[t]
 \centering
 \includegraphics[width=18cm]{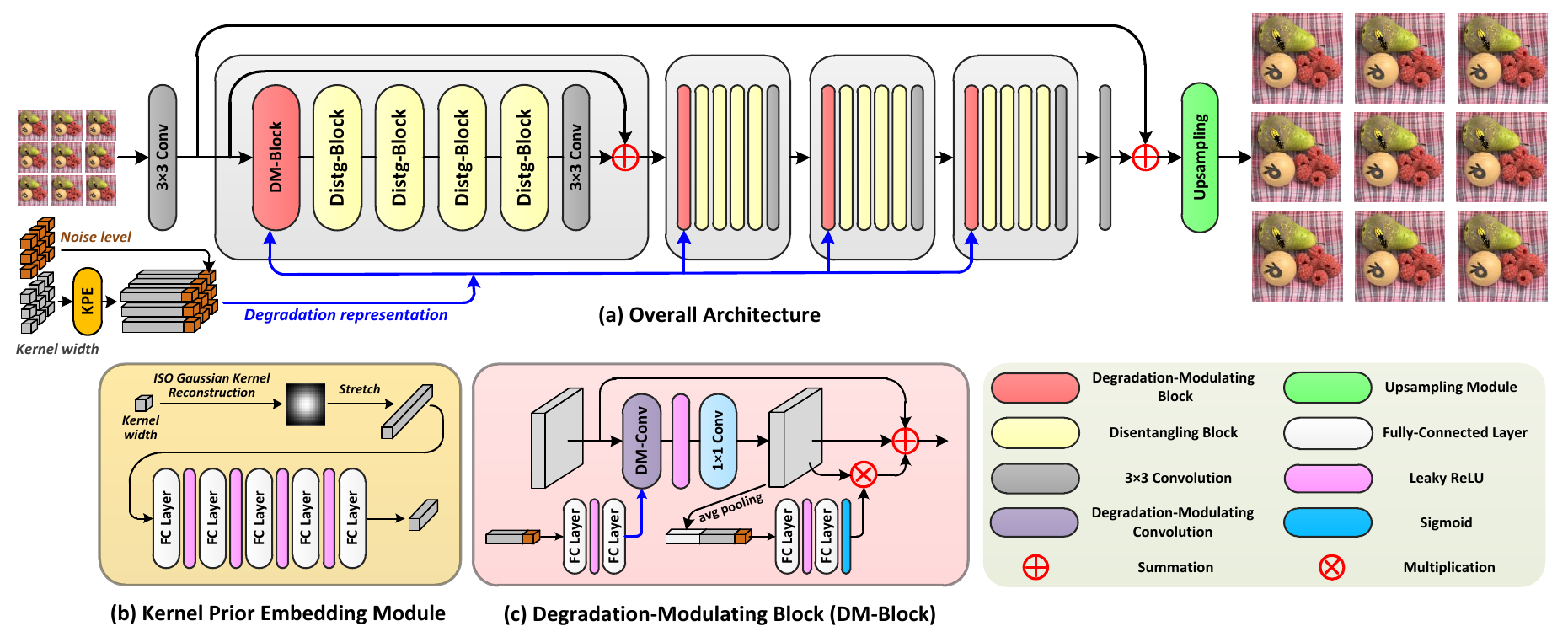}
 \caption{An overview of our LF-DMnet.}\label{fig:DMnet}
 \end{figure*}

 \subsection{Comparison to Existing Works}\label{sec:DegradationDiscuss}
 \subsubsection{Compared to Existing LF Image SR Methods}
  Compared to existing LF image SR methods \cite{resLF,ATO,LF-InterNet,LF-DFnet,MEG-Net,DistgLF,DPT,LFT} which use the bicubic downsampling approach to produce LR LF images, our method adopts a more practical degradation model (i.e., Eq.~\ref{eq:degrad}) since the blur kernel and noise level in our model can be adjusted in the training phase to enlarge the degradation space. It is shown in Section~\ref{sec:comparison-real} that our LF-DMnet trained with this degradation model can achieve promising SR performance on real LF images, which demonstrates that our proposed degradation model can well cover the real-world degradation of LF images.

 \subsubsection{Compared to More Complex Synthetic Degradation}
  It is also worth noting that several recent works for single image SR \cite{BSRGAN,RealESRGAN} designed very complex degradation models to train deep networks for real-world SR. In these methods, various kinds of blur, noise, and downsampling schemes were considered, and the order of these degradation elements (also including JPEG compression) were randomly shuffled to cover as much real-world degradation as possible. Although these methods \cite{BSRGAN,RealESRGAN} achieve favorable visual performance on real-world images, we do not consider designing such a complex degradation model in this paper because of the following three reasons. \textbf{\textit{First}}, single images are generally captured by various cameras and transmitted multiple times on internet, and thus go through complex and high-order degradation \cite{RealESRGAN}. In contrast, LF images are captured by a few kinds of imaging devices (e.g., Lytro or RayTrix), and saved to specific file formats that do not go through JPEG compression. Consequently, the degradation space of LF images is smaller than that of single images. \textbf{\textit{Second}}, abundant high-quality HR images and diverse scenarios are required to train a network to fit such complex degradation. Networks in \cite{BSRGAN,RealESRGAN} were trained on multiple large-scale single image datasets \cite{DIV2K,Flickr2K,wang2018recovering,ma2016waterloo} with thousands of high-quality HR images. In contrast, publicly available high-quality LF datasets are limited in amount, spatial resolution and scene diversity. Consequently, it is difficult for an LF image SR network to learn such complex degradation with insufficient training samples. \textbf{\textit{Third}}, as the first work to address LF image SR with multiple degradation, we aim to demonstrate the importance of degradation modulation to LF image SR, and propose a simple yet effective solution to this problem. Consequently, we do not make our degradation model over-complex.

 \section{Network Architecture}\label{sec:Network}

 \subsection{Overview}
 Based on the degradation model in Eq.~\ref{eq:degrad}, we develop a degradation-modulating network (LF-DMnet) that can super-resolve LF images with various degradation. An overview of our LF-DMnet is shown in Fig.~\ref{fig:DMnet}(a). Given an array of LR SAIs and their corresponding degradation (i.e., kernel width and noise level of each view), our LF-DMnet sequentially performs kernel prior embedding, degradation-modulated feature extraction, and upsampling. Following \cite{DistgLF}, we build our network by cascading four residual groups. In each residual group, a degradation-modulating block (DM-Block) is designed to process features according to the degradation, and four disentangling blocks (Distg-Blocks) are used to achieve spatial-angular information incorporation. The final output of our network is an array of HR LF images. Note that, since most LF image SR methods \cite{resLF,LFSSR,ATO,LF-InterNet,LF-DFnet,DistgLF,LFT} use SAIs distributed in a square array as their inputs, in this paper, we follow these methods and set $U=V=A$, where $A$ denotes the angular resolution. In the following subsections, we will introduce the details of our network design.

 \subsection{Kernel Prior Embedding}
 Handling image SR with multi-degradation is more challenging than handling that with bicubic downsampling only, since the solution space of the former one is much larger than the latter one. In such case, incorporating kernel priors into the SR process can constrain the solution space to a mainfold and thus reduce the ill-posedness of the SR process \cite{FKP}. Since only isotropic Gaussian kernel (with different kernel widths) is considered in our method, we designed a kernel prior embedding (KPE) module to fully incorporate the kernel prior into the SR process.

 In the KPE module, the isotropic Gaussian kernel $k \in \mathbb{R}^{21 \times 21}$ is first reconstructed according to the input kernel width (i.e., the only undetermined coefficient). The reconstructed kernel is then stretched into a 1D tensor $\mathbf{v}_{k} \in \mathbb{R}^{441 \times 1}$ and fed to a multi-layer perception (MLP) unit with five fully-connected (FC) layers to learn the internal characteristics. The output of the MLP is a compact blur representation with reduced dimensionality, i.e., $\mathbf{v}_{blur} \in \mathbb{R}^{15 \times 1}$. Finally, the generated blur representation is concatenated with the noise level to produce the final degradation representation $\mathbf{v}_{dg} \in \mathbb{R}^{16 \times 1}$. It is demonstrated in Section~\ref{sec:ablation} that the proposed KPE module is beneficial to the SR performance.

 \subsection{Degradation-Modulating Block}\label{sec:DM-Block}

 Degradation-modulating block (DM-Block) is designed to process image features based on the given degradation. To achieve this goal, a simple and straightforward scheme is to concatenate degradation representation with image features and fuse them via convolutions \cite{SRMD,UDVD}. However, as demonstrated in several recent works \cite{IKC,DASR}, directly convolving image features with degradation representations can cause interference since there is a domain gap between these two kinds of representations. Motivated by the fact that images with different degradation are generated by convolving the original high-quality image using isotropic Gaussian kernel with different kernel widths, in this paper, we design a degradation-modulating convolution (DM-Conv) whose kernels are dynamically generated according to the input degradation representation.

 Specifically, in each DM-Block, the degradation representation $\mathbf{v}_\textit{dg}$ is firstly fed to two FC layers to produce a convolutional kernel $\mathbf{w}$ (with a size of 3$\times$3$\times$64 in this paper). Then, the input feature $\mathcal{F}_\textit{input}$ is processed with DM-Conv (using $\mathbf{w}$) and another $1\times 1$ convolution to generate $\mathcal{F}^\textit{spa}_\textit{mod}$.
 Note that, we follow \cite{DASR} to design our DM-Conv as a depth-wise dynamic convolution, and use a channel attention layer to re-weight the output features based on the statistics of both image feature (produced by performing average pooling on $\mathcal{F}^\textit{spa}_\textit{mod}$) and degradation representation.  Specifically, the degradation representation $\mathbf{v}_\textit{dg}$ is passed to another two FC layers and a \textit{Sigmoid} activation layer to generate channel-wise modulation coefficients, which are then used to rescale different channels of $\mathcal{F}^\textit{spa}_\textit{mod}$, resulting in $\mathcal{F}_\textit{mod}$. Finally, $\mathcal{F}_\textit{mod}$ is summed up with $\mathcal{F}_\textit{input}$ and $\mathcal{F}^\textit{spa}_\textit{mod}$ to produce the output of our DM-Block. It is demonstrated in Section~\ref{sec:ablation} that our method benefits from DM-Conv and degradation-modulating channel attention, and can well handle LF images with various degradation.

  \begin{figure}[t]
 \centering
 \includegraphics[width=8.5cm]{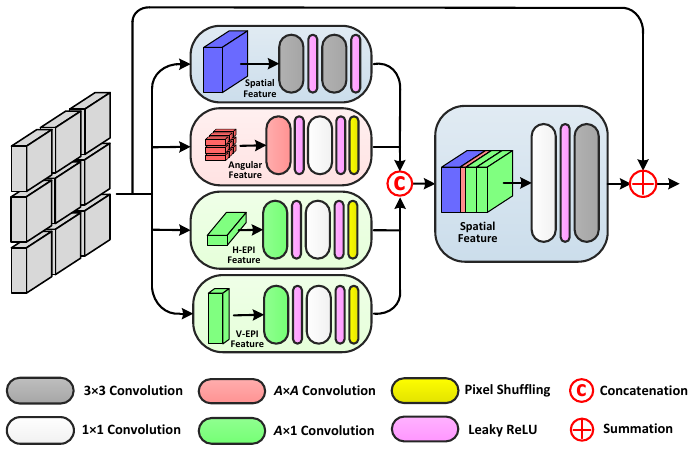}
 \caption{The architecture of our modified Distg-Block.}\label{fig:DistgBlock}
 \end{figure}

 \subsection{Disentangling Block}
 Although the proposed DM-Block can handle input images with various degradation, it processes image features of different views separately without considering the inter-view correlation. Since information both within a single view and among different views is beneficial to the performance of LF image SR, in this paper, we modify the disentangling block (Distg-Block) \cite{DistgLF} to incorporate multi-dimensional information for LF image SR.

 Different from the Distg-Block in \cite{DistgLF} where a series of specifically designed convolutions (i.e., spatial, angular and epipolar feature extractors) are applied to a single macro-pixel image (MacPI) feature, in this paper, we organize LF features into different shapes and apply plain convolutions to the reshaped features. Our modified approach is equivalent to the original design but is more simple and generic. Specifically, considering both batch and channel dimensions, the input feature of our Distg-Block can be denoted by $\mathcal{F}^{6D}_{in} \in \mathbb{R}^{B\times U\times V \times C \times H\times W}$, where $B$, $C$, $H$, $W$ represent batch, channel, height and width, respectively, and $U$$=$$V$$=$$A$ represent angular resolution. As shown in Fig.~\ref{fig:DistgBlock}, our Distg-Block has a spatial branch, an angular branch and two EPI branches (i.e., horizontal and vertical). In each branch, the input feature is reshaped into a 4D feature and then convolved by several 2D convolutions to achieve intra and inter view information incorporation.

 By adopting Distg-Blocks, our method can incorporate the beneficial spatial and angular information from the input LF to achieve state-of-the-art SR performance. The effectiveness of the Distg-Block for multi-degraded LF image SR is validated in Section~\ref{sec:ablation}.

\begin{table*}[t]
	\renewcommand\arraystretch{1.1}
	\caption{PSNR and SSIM results achieved by different methods on the HCInew \cite{HCInew}, HCIold \cite{HCIold} and STFgantry \cite{STFgantry} datasets under synthetic degradation (with different blur kernel widths and noise levels) for 4$\times$ SR. Note that, the degradation degenerates to the bicubic downsampling degradation when kernel width and noise level equal to 0. Best results are in \textbf{bold faces} and the second best results are \underline{underlined}.}\label{tab:quantitative}
	\begin{center}
		\scriptsize
		\setlength{\tabcolsep}{0.9mm}{
		\begin{tabular}{|l|c|cccc|cccc|cccc|}
			\hline
			{Method} & {Kernel} &			\multicolumn{4}{c|}{HCInew}  &
			\multicolumn{4}{c|}{HCIold} & \multicolumn{4}{c|}{STFgantry}
			\tabularnewline
			\hline
			\hline
			\multicolumn{2}{|l|}{\multirow{1}{*}{Noise level}}
			& $0$ & $15$ & $50$ & $90$
			& $0$ & $15$ & $50$ & $90$
			& $0$ & $15$ & $50$ & $90$
			\tabularnewline
			\hline
			\hline
			Bicubic & \multirow{8}{*}{$0$}
			& 27.71$/$.852 & 25.90$/$.789 & 19.53$/$.492 & 14.93$/$.276 
			& 32.58$/$.934 & 28.55$/$.857 & 20.05$/$.501 & 15.12$/$.257 
			& 26.09$/$.845 & 24.68$/$.789 & 19.18$/$.516 & 14.80$/$.304
			\tabularnewline					
			DistgSSR &
			& \underline{31.38}$/$\textbf{.922} & 24.88$/$.722 & 15.59$/$.284 & 10.24$/$.118
			& \underline{37.56}$/$\underline{.973} & 26.17$/$.751 & 15.43$/$.256 & 10.10$/$.092
			& \underline{31.66}$/$\underline{.953} & 24.37$/$.754 & 15.53$/$.319 & 10.24$/$.141
			\tabularnewline
			LFT &
			& \textbf{31.43}$/$\underline{.921} & 24.99$/$.729 & 15.89$/$.279 & 10.93$/$.107
			& \textbf{37.63}$/$\textbf{.974} & 26.48$/$.765 & 15.96$/$.258 & 10.91$/$.081
			& \textbf{31.80}$/$\textbf{.954} & 24.39$/$.758 & 15.74$/$.317 & 10.93$/$.136
			\tabularnewline
			SRMD &
			& 29.55$/$.886 & \underline{27.88}$/$.851 & \underline{25.37}$/$\underline{.806} & \underline{23.76}$/$\underline{.780}
			& 35.04$/$.953 & \underline{31.56}$/$\underline{.919} & \underline{28.26}$/$\underline{.883} & \underline{26.40}$/$\underline{.865}  
			& 28.85$/$.911 & \underline{26.73}$/$\underline{.869} & \underline{23.60}$/$\underline{.795} & \underline{21.79}$/$\underline{.747}	
			\tabularnewline
			DASR &
			& 29.31$/$.886 & 27.78$/$\underline{.852} & 24.10$/$.785 & nan$/$nan
			& 34.54$/$.950 & 31.45$/$\underline{.919} & 22.70$/$.829 & nan$/$nan
			& 26.99$/$.897 & 26.07$/$.866 & 21.92$/$.768 & nan$/$nan
			\tabularnewline
			BSRNet &
			& 28.42$/$.865 & 24.98$/$.831 & 19.32$/$.748 & 13.75$/$.631
			& 32.73$/$.933 & 28.22$/$.895 & 17.97$/$.748 & 13.47$/$.573 
			& 26.55$/$.880 & 22.55$/$.829 & 17.46$/$.714 & 14.39$/$.618
			\tabularnewline
			Real-ESRNet &
			& 28.05$/$.862 & 26.99$/$.839 & 23.65$/$.789 & 18.75$/$.728 
			& 31.80$/$.931 & 30.11$/$.905 & 24.14$/$.842 & 17.95$/$.734 
			& 24.78$/$.871 & 24.51$/$.850 & 19.45$/$.754 & 16.31$/$.690 
			\tabularnewline
			Ours &
			& 30.43$/$.907 & \textbf{29.55}$/$\textbf{.886} & \textbf{28.23}$/$\textbf{.859} & \textbf{27.21}$/$\textbf{.839}
			& 36.44$/$.967 & \textbf{34.63}$/$\textbf{.951} & \textbf{32.42}$/$\textbf{.929} & \textbf{30.88}$/$\textbf{.912} 
			& 29.77$/$.932 & \textbf{28.62}$/$\textbf{.912} & \textbf{26.99}$/$\textbf{.878} & \textbf{25.74}$/$\textbf{.848}
			\tabularnewline
			\hline
			\hline
			Bicubic & \multirow{8}{*}{$1.5$}
			& 27.02$/$.836 & 25.42$/$.773 & 19.41$/$.478 & 14.89$/$.266
			& 31.63$/$.923 & 28.16$/$.846 & 19.99$/$.491 & 15.10$/$.252 
			& 25.15$/$.821 & 24.00$/$.764 & 18.96$/$.493 & 14.72$/$.287
			\tabularnewline
			DistgSSR &
			& 28.60$/$.876 & 24.46$/$.699 & 15.60$/$.273 & 10.23$/$.112 
			& 33.64$/$.949 & 25.97$/$.739 & 15.43$/$.251 & 10.10$/$.089 
			& 27.16$/$.883 & 23.59$/$.714 & 15.57$/$.302 & 10.27$/$.131 
			\tabularnewline
			LFT &
			& 28.57$/$.875 & 24.60$/$.708 & 15.89$/$.268 & 10.93$/$.101 
			& 33.62$/$.949 & 26.25$/$.753 & 15.96$/$.252 & 10.92$/$.079 
			& 27.13$/$.882 & 23.69$/$.721 & 15.70$/$.295 & 10.93$/$.124 
			\tabularnewline
			SRMD &
			& \underline{29.58}$/$\underline{.886} & \underline{27.39}$/$\underline{.840} & \underline{25.01}$/$\underline{.798} & \underline{23.50}$/$\underline{.774}
			& \underline{35.00}$/$\underline{.953} & \underline{31.02}$/$\underline{.912} & \underline{27.94}$/$\underline{.879} & \underline{26.20}$/$\underline{.862}	 
			& \underline{28.87}$/$\underline{.910} & \underline{26.05}$/$\underline{.851} & \underline{23.06}$/$\underline{.776} &  \underline{21.40}$/$\underline{.732} 
			\tabularnewline
			DASR &
			& 29.46$/$.884 & 27.34$/$\underline{.840} & 24.09$/$.781 & nan$/$nan
			& 34.87$/$.952 & 30.95$/$.911 & 23.44$/$.831 & nan$/$nan
			& 27.83$/$.902 & 25.84$/$.850 & 21.95$/$.755 & nan$/$nan
			\tabularnewline
			BSRNet &
			& 28.38$/$.861 & 24.79$/$.824 & 19.36$/$.746 & 13.80$/$.632 
			& 32.77$/$.932 & 28.11$/$.892 & 18.00$/$.749 & 13.48$/$.574 
			& 26.67$/$.877 & 22.34$/$.815 & 17.39$/$.706 & 14.46$/$.618 
			\tabularnewline
			Real-ESRNet &
			& 28.17$/$.862 & 26.68$/$.830 & 23.50$/$.783 & 18.65$/$.724 
			& 32.11$/$.932 & 29.85$/$.900 & 24.13$/$.840 & 17.91$/$.731
			& 25.18$/$.872 & 24.30$/$.834 & 19.41$/$.741 & 16.22$/$.682
			\tabularnewline
			Ours &
			& \textbf{30.15}$/$\textbf{.900} & \textbf{28.98}$/$\textbf{.872} & \textbf{27.65}$/$\textbf{.845} & \textbf{26.70}$/$\textbf{.826}
			& \textbf{36.10}$/$\textbf{.963} & \textbf{33.87}$/$\textbf{.942} & \textbf{31.81}$/$\textbf{.920} & \textbf{30.36}$/$\textbf{.905}
			& \textbf{29.47}$/$\textbf{.924} & \textbf{27.91}$/$\textbf{.894} & \textbf{26.25}$/$\textbf{.857} & \textbf{25.05}$/$\textbf{.829} 
			\tabularnewline
			\hline
			\hline
			Bicubic & \multirow{8}{*}{$3$}
			& 25.52$/$.803 & 24.32$/$.741 & 19.09$/$.454 & 14.77$/$.250 
			& 29.59$/$.898 & 27.12$/$.822 & 19.82$/$.476 & 15.04$/$.243 
			& 23.21$/$.766 & 22.45$/$.711 & 18.41$/$.450 & 14.50$/$.258 
			\tabularnewline
			DistgSSR &
			& 25.79$/$.811 & 23.30$/$.656 & 15.47$/$.254 & 10.21$/$.104 
			& 29.92$/$.904 & 25.19$/$.710 & 15.38$/$.241 & 10.10$/$.086 
			& 23.55$/$.780 & 21.83$/$.639 & 15.32$/$.265 & 10.19$/$.113 
			\tabularnewline
			LFT &
			& 25.73$/$.810 & 23.41$/$.665 & 15.75$/$.248 & 10.87$/$.090
			& 29.83$/$.904 & 25.42$/$.724 & 15.91$/$.242 & 10.89$/$.075 
			& 23.47$/$.779 & 21.89$/$.647 & 15.43$/$.257 & 10.81$/$.104 
			\tabularnewline
			SRMD &
			& \underline{29.20}$/$\underline{.876} & \underline{26.32}$/$\underline{.816} & \underline{24.30}$/$\underline{.782} & \underline{23.04}$/$\underline{.763} 
			& \underline{34.39}$/$\underline{.948} & \underline{29.87}$/$\underline{.896} & \underline{27.36}$/$\underline{.871} & \underline{25.79}$/$\underline{.857}
			& \underline{28.29}$/$\underline{.898} & \underline{24.51}$/$\underline{.807} & \underline{22.08}$/$\underline{.742} & \underline{20.80}$/$\underline{.710} 
			\tabularnewline	
			DASR &
			& 28.62$/$.867 & 26.26$/$.815 & 23.70$/$.767 & nan$/$nan
			& 33.72$/$.942 & 29.82$/$\underline{.896} & 23.75$/$.829 & nan$/$nan
			& 27.71$/$.887 & 24.48$/$\underline{.807} & 21.50$/$.723 & nan$/$nan
			\tabularnewline	
			BSRNet &
			& 27.60$/$.843 & 24.13$/$.804 & 19.33$/$.740 & 13.84$/$.633
			& 31.96$/$.921 & 27.72$/$.883 & 18.05$/$.750 & 13.51$/$.576
			& 26.05$/$.849 & 21.62$/$.776 & 17.23$/$.691 & 14.50$/$.618
			\tabularnewline	
			Real-ESRNet &
			& 27.33$/$.845 & 25.67$/$.807 & 23.04$/$.769 & 18.54$/$.716
			& 31.45$/$.919 & 29.11$/$.889 & 23.93$/$.835 & 17.84$/$.728
			& 25.24$/$.856 & 23.35$/$.789 & 19.14$/$.712 & 16.08$/$.670 
			\tabularnewline						
			Ours &
			& \textbf{29.43}$/$\textbf{.884} & \textbf{27.76}$/$\textbf{.845} & \textbf{26.54}$/$\textbf{.821} & \textbf{25.74}/\textbf{.806} 
			& \textbf{35.10}$/$\textbf{.955} & \textbf{32.34}$/$\textbf{.924} & \textbf{30.54}$/$\textbf{.904} & \textbf{29.43}/\textbf{.892}
			& \textbf{28.51}$/$\textbf{.904} & \textbf{26.22}$/$\textbf{.853} & \textbf{24.72}$/$\textbf{.814} & \textbf{23.74}/\textbf{.788} 
			\tabularnewline
			\hline
			\hline
			Bicubic & \multirow{8}{*}{$4.5$}
			& 24.36$/$.779 & 23.41$/$.718 & 18.79$/$.438 & 14.65$/$.240
			& 28.05$/$.879 & 26.19$/$.803 & 19.63$/$.465 & 14.97$/$.237
			& 21.80$/$.725 & 21.26$/$.672 & 17.90$/$.420 & 14.29$/$.239
			\tabularnewline
			DistgSSR &
			& 24.38$/$.781 & 22.48$/$.631 & 15.33$/$.242 & 10.16$/$.099
			& 28.08$/$.880 & 24.50$/$.690 & 15.31$/$.235 & 10.08$/$.084
			& 21.83$/$.728 & 20.67$/$.595 & 15.04$/$.243 & 10.12$/$.103
			\tabularnewline
			LFT &
			& 24.39$/$.781 & 22.57$/$.640 & 15.61$/$.237 & 10.82$/$.085
			& 28.08$/$.880 & 24.71$/$.705 & 15.84$/$.235 & 10.85$/$.072 
			& 21.84$/$.728 & 20.72$/$.602 & 15.14$/$.233 & 10.72$/$.093 
			\tabularnewline
			SRMD &
			& \underline{26.32}$/$\underline{.818} & \underline{25.09}$/$\underline{.792} & \underline{23.65}$/$\underline{.769} & \underline{22.62}$/$\underline{.756}
			& \underline{30.62}$/$\underline{.908} & \underline{28.61}$/$\underline{.882} & \underline{26.66}$/$\underline{.864} & \underline{25.38}$/$\underline{.853}
			& \underline{24.34}$/$\underline{.780} & \underline{22.80}$/$\underline{.753} & \underline{21.28}$/$\underline{.716} & \underline{20.37}$/$\underline{.697} 
			\tabularnewline	
			DASR &
			& 25.34$/$.799 & 24.89$/$.788 & 23.11$/$.755 & nan$/$nan
			& 29.33$/$.895 & 28.39$/$.880 & 23.94$/$.827 & nan$/$nan
			& 22.99$/$.761 & 22.65$/$.749 & 20.76$/$.697 & nan$/$nan
			\tabularnewline	
			BSRNet &
			& 26.31$/$.816 & 23.40$/$.784 & 19.26$/$.734 & 13.85$/$.634
			& 30.35$/$.902 & 27.10$/$.874 & 18.03$/$.749 & 13.51$/$.577 
			& 24.23$/$.795 & 20.83$/$.738 & 17.06$/$.680 & 14.55$/$.618
			\tabularnewline	
			Real-ESRNet &
			& 26.28$/$.816 & 24.69$/$.787 & 22.55$/$.758 & 18.45$/$.711
			& 30.04$/$.900 & 28.08$/$.878 & 23.66$/$.830 & 17.77$/$.726 
			& 23.97$/$.810 & 22.17$/$.743 & 18.90$/$.693 & 15.94$/$.660
			\tabularnewline						
			Ours &
			& \textbf{28.00}$/$\textbf{.854} & \textbf{26.55}$/$\textbf{.820} & \textbf{25.58}$/$\textbf{.801} & \textbf{24.84}$/$\textbf{.788} 
			& \textbf{33.39}$/$\textbf{.937} & \textbf{30.88}$/$\textbf{.906} & \textbf{29.45}$/$\textbf{.890} & \textbf{28.52}$/$\textbf{.880}
			& \textbf{26.59}$/$\textbf{.860} & \textbf{24.64}$/$\textbf{.808} & \textbf{23.30}$/$\textbf{.771} & \textbf{22.47}$/$\textbf{.748} 
			\tabularnewline
			\hline						
			\end{tabular}}
		\vspace{-0.05cm}
	\end{center}
	\leftline{~~Note: 1) For the methods in \cite{BSRGAN,RealESRGAN}, we used the models trained with a pixel-wise L1 loss (i.e., BSRNet and Real-ESRNet) for comparison since they}
	\leftline{~~can achieve higher PSNR and SSIM values as compared to their GAN-based version (i.e., BSRGAN and Real-ESRGAN). 2) SRMD and our LF-DMnet}
	\leftline{~~are non-blind SR methods while DASR, BSRNet and Real-ESRNet are blind SR methods.}
\end{table*}

\subsection{Discussion on the Non-Blind SR Setting}
 Recent single image SR methods \cite{IKC,DASR,FKP,MADNet} generally adopt the blind SR settings, i.e., the ``groundtruth'' degradation is unknown for the SR networks. That is because, compared to non-blind SR methods \cite{SRMD,UDVD,USRnet,DPSR} where the degradation is also required as the input, blind SR is more practical since the real-world degradation is generally difficult to obtain.

 However, in this paper, we adopt the non-blind SR settings as in \cite{SRMD}, and take both degraded LF images and their degradation (blur kernel width and noise level) as inputs of our network. Reasons are in three folds.
 \textbf{\textit{First}}, performing non-blind SR helps us to better investigate the impact of input degradation to the SR performance, which has not been studied in LF image SR. Since the kernel width and noise level are independently fed to our network, performing non-blind SR can help us decouple different degradation elements and investigate their influence respectively, as demonstrated in Section~\ref{sec:mismatch}.
 \textbf{\textit{Second}}, performing non-blind SR helps us to explore the upper bound of blind SR because the groundtruth degradation information can be used as an accurate prior in non-blind SR. As the first work to achieve LF image SR with multi-degradation, one of the major contributions of this paper is to break the limitation of single fixed degradation and show the great potential and practical values of multi-degraded LF image SR. To this end, non-blind SR is purer and more suitable than blind SR.
\textbf{\textit{Third}}, since the proposed degradation model has only two under-determined coefficients, we can easily find a proper input degradation by observing the super-resolved images to correct the input degradation, or adopting a grid search strategy \cite{SRMD} to traverse kernel widths and noise levels in a reasonable range, as described in Section~\ref{sec:mismatch-real}.

\begin{figure*}[t]
 \centering
 \includegraphics[width=18cm]{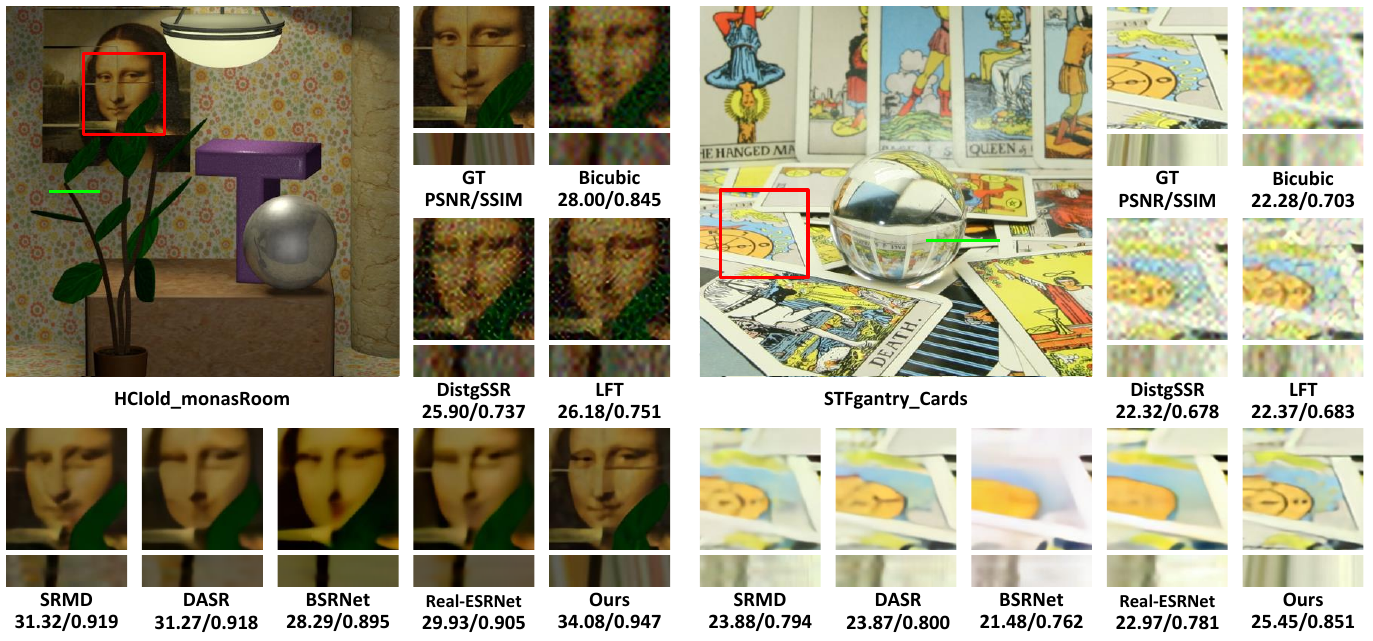}
 \caption{Visual results achieved by different methods on synthetically degraded LFs ($\textit{kernel width}=1.5$, $\textit{noise level}=15$) for 4$\times$SR. The super-resolved center view images and horizontal EPIs are shown. The PSNR and SSIM scores on the presented scenes are reported below the zoom-in regions.}\label{fig:visual-syn}
 \end{figure*}

\section{Experiments}\label{sec:Experiment}
In this section, we first introduce the datasets and implementation details, then compare our network to several state-of-the-art SR methods. Finally, we conduct ablation studies to investigate our design choices and further analyze the impact of the input kernel widths and noise levels.

\subsection{Datasets and Implementation Details}
Our method was trained and validated on synthetically degraded LFs generated according to Eq.~\ref{eq:degrad}, and further tested on real LFs captured by Lytro Illum and Raytrix cameras. For training and validation, three public LF datasets including HCInew \cite{HCInew}, HCIold \cite{HCIold} and STFgantry \cite{STFgantry} were adopted. The division of training and validation set was kept identical to that in \cite{LF-DFnet,DistgLF,LFT,DPT}. To test the generalization capability of our method to real-world degradation,  three public LF datasets (i.e., EPFL \cite{EPFL}, INRIA \cite{INRIA} and STFlytro \cite{STFlytro}) developed with Lytro cameras and a dataset \cite{Raytrix} developed with a Raytrix camera were used as our test sets. Totally 39, 8 and 26 scenes were used for training, validation and test in this paper, respectively.

The LFs in the HCInew \cite{HCInew}, HCIold \cite{HCIold}, STFgantry \cite{STFgantry}, EPFL \cite{EPFL}, INRIA \cite{INRIA} and STFlytro \cite{STFlytro} datasets have an angular resolution of 9$\times$9, and the LFs in the Raytrix dataset \cite{Raytrix} have an angular resolution of 5$\times$5. For LFs with an angular resolution of 9$\times$9, we followed the existing works \cite{LF-InterNet, LF-DFnet, LFT, DPT, DistgLF} to use the central 5$\times$5 SAIs in our experiments. In the training phase, we cropped HR SAIs into patches of size 152$\times$152 with a stride of 32, and used the proposed degradation model to synthesize LR SAI patches of size 38$\times$38\footnote{Following \cite{BSRGAN,RealESRGAN}, we only consider 4$\times$SR in this paper.}. We followed \cite{IKC,DASR} to set the window size of the isotropic Gaussian kernel to 21$\times$21, and followed \cite{SRMD} to randomly sample the kernel width and noise level from range $[0, 4]$ and $[0, 75]$, respectively. Note that, to avoid boundary effect caused by Gaussian filtering, only central 128$\times$128 region of the HR patches and their corresponding 32$\times$32 LR patches were used for training. We performed random horizontal flipping, vertical flipping, 90-degree rotation and RGB channel shuffling to augment the training data by 48 times. Note that, the spatial and angular dimension need to be flipped or rotated jointly to maintain LF structures.

Our network was trained using the L1 loss and optimized using the Adam method \cite{Adam} with $\beta_1$=0.9, $\beta_2$=0.999 and a batch size of 8. Our LF-DMnet was implemented in PyTorch on a PC with two NVidia RTX 2080Ti GPUs. The learning rate was initially set to $2\times10^{-4}$ and decreased by a factor of 0.5 for every 3$\times$10$^4$ iterations. The training was stopped after 10$^5$ iterations.

Following \cite{SRMD,IKC,DASR}, we used PSNR and SSIM calculated on the RGB channel images as quantitative metrics for validation. To obtain the metric score (e.g., PSNR) for a dataset with $M$ scenes (each scene has an angular resolution of $A \times A$), we first calculated the metric on $A\times A$ SAIs on each scene separately, then obtained the score for each scene by averaging its $A^2$ scores, and finally obtained the score for this dataset by averaging the scores of all $M$ scenes.

\subsection{Comparisons with State-of-the-art Methods}
  In this subsection, we compare our method to the following state-of-the-art SR methods:
  \begin{itemize}
  \item \textbf{DistgSSR} \cite{DistgLF} and \textbf{LFT} \cite{LFT}: two top-performing LF image SR methods developed on the bicubic downsampling degradation;
  \item \textbf{SRMD} \cite{SRMD}: a popular non-blind single image SR method developed on isotropic Gaussian blur and Gaussian noise degradation;
  \item \textbf{DASR} \cite{DASR}: a state-of-the-art blind single image SR method developed on anisotropic Gaussian blur and Gaussian noise degradation;
  \item \textbf{BSRGAN} \cite{BSRGAN} and \textbf{Real-ESRGAN} \cite{RealESRGAN}: two recent real-world single image SR methods developed on the complex synthetic degradation.
  \end{itemize}
  Besides the aforementioned compared methods, we also include bicubic upsampling method to produce baseline results.

 \begin{figure*}[t]
 \centering
 \includegraphics[width=18cm]{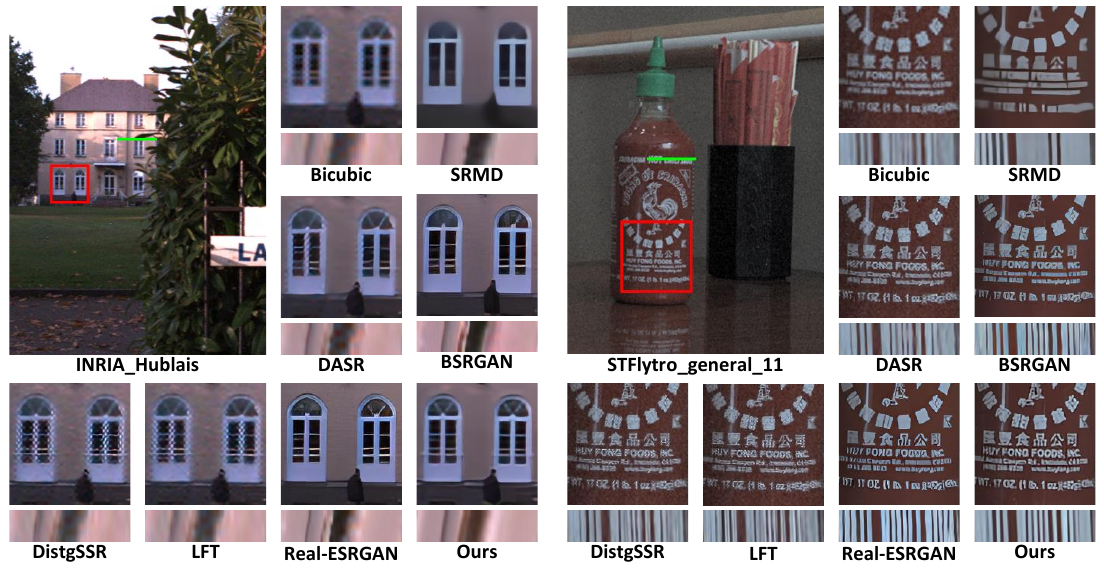}
 \caption{Visual results achieved by different methods on real LFs captured by Lytro Illum cameras for 4$\times$SR. Scenes \textit{Hublais} from the INRIA dataset \cite{INRIA} and \textit{general\_11} from the STFlytro dataset \cite{STFlytro} are used as example scenes for comparison. The super-resolved center view images and horizontal EPIs are shown. For SRMD and our method, the input blur kernel width and noise level are set to 2 and 30, respectively. Groundtruth HR images are unavailable in this case.}\label{fig:visual-real}
 \end{figure*}

 \begin{figure*}[t]
 \centering
 \includegraphics[width=18cm]{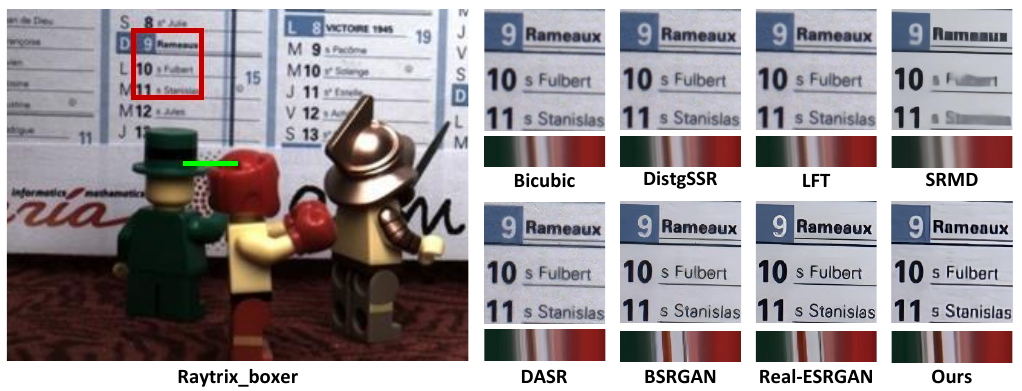}
 \caption{Visual results achieved by different methods on real LFs captured by a Raytrix camera for 4$\times$SR. Scenes \textit{boxer} from the dataset in \cite{Raytrix} is used as an example scene for comparison. The super-resolved center view images and horizontal EPIs are shown. For SRMD and our method, the input blur kernel width and noise level are set to 4 and 60, respectively. Groundtruth HR images are unavailable in this case.}\label{fig:visual-raytrix}
 \end{figure*}

 \subsubsection{Results on Synthetically Degraded LFs}\label{sec:comparison-syn}
  Table~\ref{tab:quantitative} shows the quantitative PSNR and SSIM results achieved by different methods under synthetic degradation with different blur and noise levels. It can be observed that DistgSSR and LFT produce the top-2 highest PSNR and SSIM results under the bicubic downsampling degradation (i.e., $\textit{kernel width}=0$, $\textit{noise level}=0$), but suffer from significant performance drop when the kernel width and noise level are larger than zero. This demonstrates that existing LF image SR methods trained on the noise-free bicubic downsampling degradation cannot generalize well to other degradation.

  SRMD and DASR achieves much better performance than DistgSSR and LFT on blurry and noisy scenes since these two methods are designed for multi-degraded image SR. Note that, SRMD is benefited from the input groundtruth degradation and thus slightly outperforms DASR. It can be also observed that the PSNR and SSIM values produced by BSRNet  and Real-ESRNet are lower than SRMD and DASR. That is because, the degradation space in BSRNet  and Real-ESRNet are much larger, so that the capability of these two methods in handling specific degradation is less powerful. It is worth noting that these single image SR methods only use spatial context information within single views for SR but overlook the correlations among different views, resulting in inferior SR performance and the angular inconsistency issue (see Section~\ref{sec:AngCons}).

  Compared to these state-of-the-art single and LF image SR methods, our LF-DMnet can simultaneously incorporate the complementary angular information and adapt to different degradation, and thus achieves the best PSNR and SSIM results on both in-distribution degradation and out-of-distribution (e.g., \textit{kernel width}$=$$4.5$ or \textit{noise level}$=$$90$) degradation except for the noise-free bicubic downsampling one.
  The benefits of angular information and degradation adaption are further analyzed in Section~\ref{sec:ablation}. Figure~\ref{fig:visual-syn} shows the visual results produced by different methods with blur kernel width and noise level being set to 1.5 and 15, respectively. It can be observed that our LF-DMnet can recover faithful details from the blurry and noisy input LFs.

\subsubsection{Results on Real LFs}\label{sec:comparison-real}
We test the practical values of different SR methods by directly applying them to LFs captured by Lytro and Raytrix cameras. Since the groundtruth HR images of the input LFs are unavailable, we compare the visual results produced by different methods in Figs.~\ref{fig:visual-real} and \ref{fig:visual-raytrix}. It can be observed that the image quality of the input LFs is low since the bicubicly upsampled images are blurry and noisy. DistgSSR and LFT augment the input noise and produce results with artifacts (see Fig.~\ref{fig:visual-real}) or blurring details (see Fig.~\ref{fig:visual-raytrix}). This demonstrates that methods developed on the fixed bicubic downsampling degradation cannot handle real-world degradation and thus have limited practical values.

Although SRMD, DASR, BSRGAN and Real-ESRGAN are specifically designed to handle image SR with multiple degradation, these methods do not consider inter-view correlation and ignore the beneficial angular information. Consequently, these single image SR methods suffer from noise residual (e.g., see the results of DASR in Figs.~\ref{fig:visual-real} and \ref{fig:visual-raytrix}), over-smoothness (e.g., see the results of SRMD in Figs.~\ref{fig:visual-real} and \ref{fig:visual-raytrix}), and angular inconsistency (e.g., see the results of BSRGAN and Real-ESRGAN in Fig.~\ref{fig:visual-raytrix}) issues.

Compared to existing methods, our method achieves the best SR performance on real LFs, i.e., the results produced by our method have finer details (e.g., the words and characters in scene \textit{general\_11}) and less artifacts. This demonstrates that our network trained on the proposed degradation model can effectively handle real LF image SR problem. Readers are referred to the videos\footnote{\noindent \url{https://github.com/YingqianWang/LF-DMnet/blob/main/demo_videos.md}} to view more visual SR results on real LFs.

\begin{table}
\caption{Comparisons of the number of parameters (\#Param.), FLOPs and running time for 4$\times$SR. Note that, FLOPs and running time are calculated on an input LF with an angular resolution of 5$\times$5 and a spatial resolution of 32$\times$32. PSNR and SSIM scores are averaged over 9 degradation (\textit{kernel width}$=$(0, 1.5, 3), \textit{noise level}$=$(0, 15, 50)) in Table~\ref{tab:quantitative}. Best results are in \textbf{bold faces}.}\label{tab:efficiency}
\centering
\scriptsize
\renewcommand\arraystretch{1.1}
\setlength{\tabcolsep}{0.7mm}
\begin{tabular}{|l|c|c|c|c|c|c|}
\hline
  &  \multirow{2}{*}{\#Param.}  & \multirow{2}{*}{FLOPs} & \multirow{2}{*}{Time} & \multicolumn{3}{c|}{PSNR$/$SSIM} \\
  \cline{5-7}
    & & & & HCInew & HCIold & STFgantry \\
\hline
SRMD	 \cite{SRMD}     & 1.50M & 39.76G  & 0.070s & 27.18$/$0.838 & 31.16$/$0.913 & 25.78$/$0.840 \\
DASR	 \cite{DASR}     & 5.80M & 82.03G  & 0.051s & 26.74$/$0.831 & 29.47$/$0.896 & 24.92$/$0.829 \\
BSRNet \cite{BSRGAN}	  & 16.70M & 459.6G  & 0.119s & 24.03$/$0.807 & 26.17$/$0.856 & 21.98$/$0.793 \\
Real-ESRNet	\cite{RealESRGAN}  & 16.70M & 459.6G  & 0.119s & 25.92$/$0.821 & 28.52$/$0.888 & 22.82$/$0.809 \\
DistgSSR \cite{DistgLF} & 3.53M & 65.41G  & \textbf{0.037s} & 22.79$/$0.611 & 24.97$/$0.642 & 22.06$/$0.623 \\
LFT  \cite{LFT}  & \textbf{1.11M} & \textbf{29.45G}  & 0.070s & 22.92$/$0.612 & 25.23$/$0.647 & 22.14$/$0.623 \\
LF-DMnet (ours) & 3.80M & 65.93G  & 0.039s & \textbf{28.75}$/$\textbf{0.869} & \textbf{33.69}$/$\textbf{0.939} & \textbf{27.61}$/$\textbf{0.885} \\
\hline
\end{tabular}
\end{table}

\subsubsection{Angular Consistency}\label{sec:AngCons}
Since LF image SR methods are required to preserve the LF parallax structure and generate angular-consistent HR LF images, we evaluate the angular consistency of different SR methods by visualizing their EPI slices. As shown below the zoom-in regions in Figs.~\ref{fig:visual-syn}, \ref{fig:visual-real} and \ref{fig:visual-raytrix}, our LF-DMnet can generate more straight and clear line patterns than other SR methods on both synthetic and real-world degradation, which demonstrates that the LF parallax structure is well preserved by our method. Readers can refer to this video\footnote{\url{https://wyqdatabase.s3.us-west-1.amazonaws.com/LF-DMnet.mp4}} for a visual comparison of angular consistency.

\begin{table*}
	\caption{PSNR values achieved by LF-DMnet and its variants for $4\times$SR. Here, we report the number of parameters (\#Params.), FLOPs and running time of each model for efficiency evaluation.}\label{tab:ablation}
	\centering
	\scriptsize
	\renewcommand\arraystretch{1.1}
	\setlength{\tabcolsep}{1.5mm}
	\begin{tabular}{|c|c|c|c|c|c|c|c|ccc|ccc|ccc|c|}
		\hline
		\multirow{2}*{Model} & \multirow{2}*{DM-Conv} & \multirow{2}*{DM-CA} & \multirow{2}*{KPE} & \multirow{2}*{Ang} & \multirow{2}*{\#Params.} & \multirow{2}*{FLOPs} & \multirow{2}*{Time} & \multicolumn{3}{c|}{Noise = 0} & \multicolumn{3}{c|}{Noise = 15} & \multicolumn{3}{c|}{Noise = 50} & \multirow{2}*{Average} \\
		\cline{9-17}
		& & & & & &  & & $\sigma_b$=0 & $\sigma_b$=1.5 & $\sigma_b$=3 & $\sigma_b$=0 & $\sigma_b$=1.5 & $\sigma_b$=3 & $\sigma_b$=0 & $\sigma_b$=1.5 & $\sigma_b$=3 & \\
		\hline
		Model 1  & & & & & 3.94M & 100.9G & 0.029s &
		27.87 & 27.72 & 26.97 & 26.40 & 25.71 & 24.21 & 23.40 & 22.85 & 21.87 & 25.22 \\
		\hline
		Model 2 & & & & $\checkmark$ & 3.77M & 69.18G & 0.038s &
		28.94 & 28.28 & 27.21 & 28.14 & 27.31 & 25.82 & 26.82 & 25.99 & 24.48 & 26.00 \\
		\hline
		Model 3 & $\checkmark$ & $\checkmark$ & $\checkmark$ &  & 4.01M & 97.61G & 0.030s &
		27.95 & 28.00 & 27.49 & 26.43 & 25.79 & 24.35 & 23.43 & 22.92 & 21.90 & 25.36 \\
		\hline
		Model 4 & $\checkmark$ & $\checkmark$ & & $\checkmark$ & 3.77M & 65.93G & 0.038s &
		29.36 & 28.91 & 27.90 & 28.56 & 27.77 & 26.20 & 26.95 & 26.14 & 24.70 & 27.39 \\
		\hline
		Model 5 & $\checkmark$ & & $\checkmark$ & $\checkmark$ & 3.79M & 65.93G & 0.037s &
		29.15 & 29.29 & 28.25 & 28.33 & 27.85 & 26.19 & 26.84 & 26.16 & 24.63 & 27.41 \\
		\hline
		
		\textbf{LF-DMnet} & $\checkmark$ & $\checkmark$ & $\checkmark$ & $\checkmark$ & 3.80M & 65.93G & 0.039s &
		\textbf{29.77} & \textbf{29.47} & \textbf{28.51} & \textbf{28.62} & \textbf{27.91} & \textbf{26.22} & \textbf{26.99} & \textbf{26.25} & \textbf{24.72} & \textbf{27.61} \\
		\hline
	\end{tabular}
\end{table*}

\subsubsection{Efficiency}\label{sec:efficiency}
We compare our LF-DMnet to existing SR methods in terms of the number of parameters, FLOPs and running time. As shown in Table~\ref{tab:efficiency}, our LF-DMnet has a moderate model size which is slightly larger than DistgSSR due to the additional kernel prior embedding branch and the degradation-modulating blocks. Note that, these additional 0.27M parameters only result in a 0.52G and 0.002s increase in FLOPs and running time, respectively. Compared to DASR, BSRNet (i.e., BSRGAN) and Real-ESRNet (i.e., Real-ESRGAN), our method has significantly smaller model size, lower FLOPs and shorter running time. These results demonstrate the efficiency of our method.

\begin{figure*}[t]
 \centering
 \includegraphics[width=18cm]{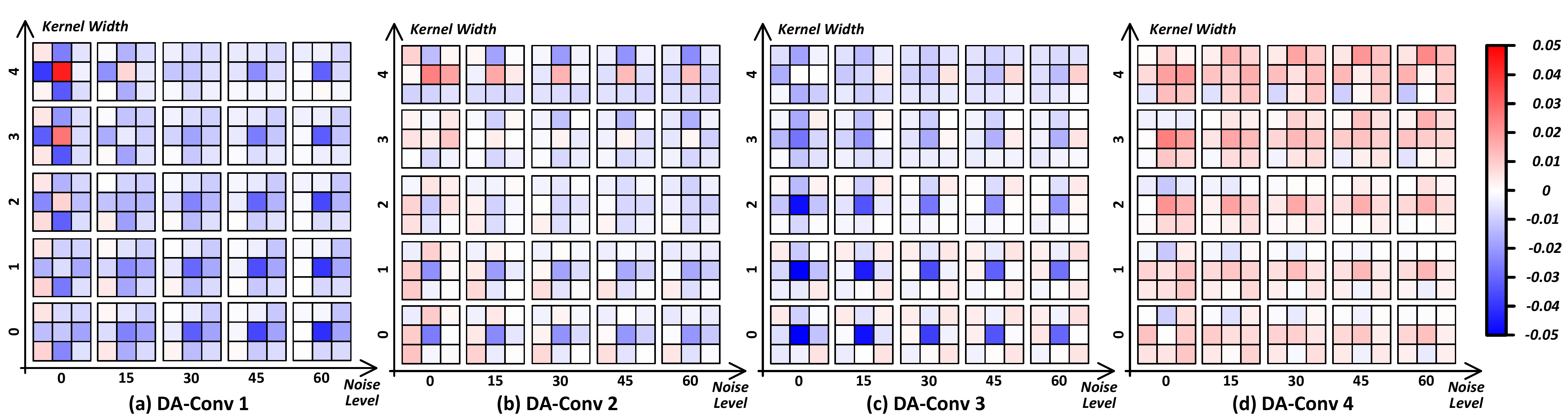}
 \caption{Kernel visualization of our degradation-modulating convolutions (DM-Convs) with different input blur and noise levels.}\label{fig:visual-kernel}
 \end{figure*}

\subsection{Ablation Study}\label{sec:ablation}
In this subsection, we investigate the effectiveness of our proposed modules and design choices by comparing our LF-DMnet with the following variants.
\begin{itemize}
\item \textbf{Model 1}: We introduce a baseline model by removing the DM-Block and the angular and EPI branches in the Distg-Block. \textit{Consequently, this variant is equivalent to a plain single image SR network that neither performs degradation modulation nor incorporates angular information.} Note that, we increase the number of convolution layers in this variant to make its model size not smaller than our LF-DMnet.
\item \textbf{Model 2}: We investigate the effectiveness of our DM-Conv by replacing it with a depth-wise 3$\times$3 convolution and a vanilla 3$\times$3 convolution. Distg-Block is maintained in this variant to incorporate angular information. Note that, the KPE module is also removed since vanilla convolutions do not take degradation as their input. \textit{This variant can be considered as an LF image SR method without degradation modulation (e.g., DistgSSR) retrained on our proposed degradation model.}
\item \textbf{Model 3}: In this variant, we remove the angular and EPI branches in the Distg-Block and adopt the same strategy as in Model 1 to make the model size of this variant not smaller than our LF-DMnet.  \textit{Since this model only incorporates intra-view information to achieve degradation-modulated SR, it can be considered as a non-blind single image SR method}, and the benefits of the angular information to real-world LF image SR can be validated.
\item \textbf{Model 4}: We modify the KPE module in this variant to investigate the effectiveness of kernel prior embedding. Specifically, we do not perform isotropic Gaussian kernel reconstruction but directly fed the blur kernel width to a five-layer MLP to generate the blur degradation representation. Consequently, the isotropic Gaussian kernel prior cannot be incorporated by this variant.
\item \textbf{Model 5}: In this variant, we remove the degradation-modulating channel attention layer (i.e., DM-CA) from the DM-Block to investigate the benefits of channel-wise degradation modulation.
\end{itemize}

\begin{figure*}[t]
 \centering
 \includegraphics[width=18cm]{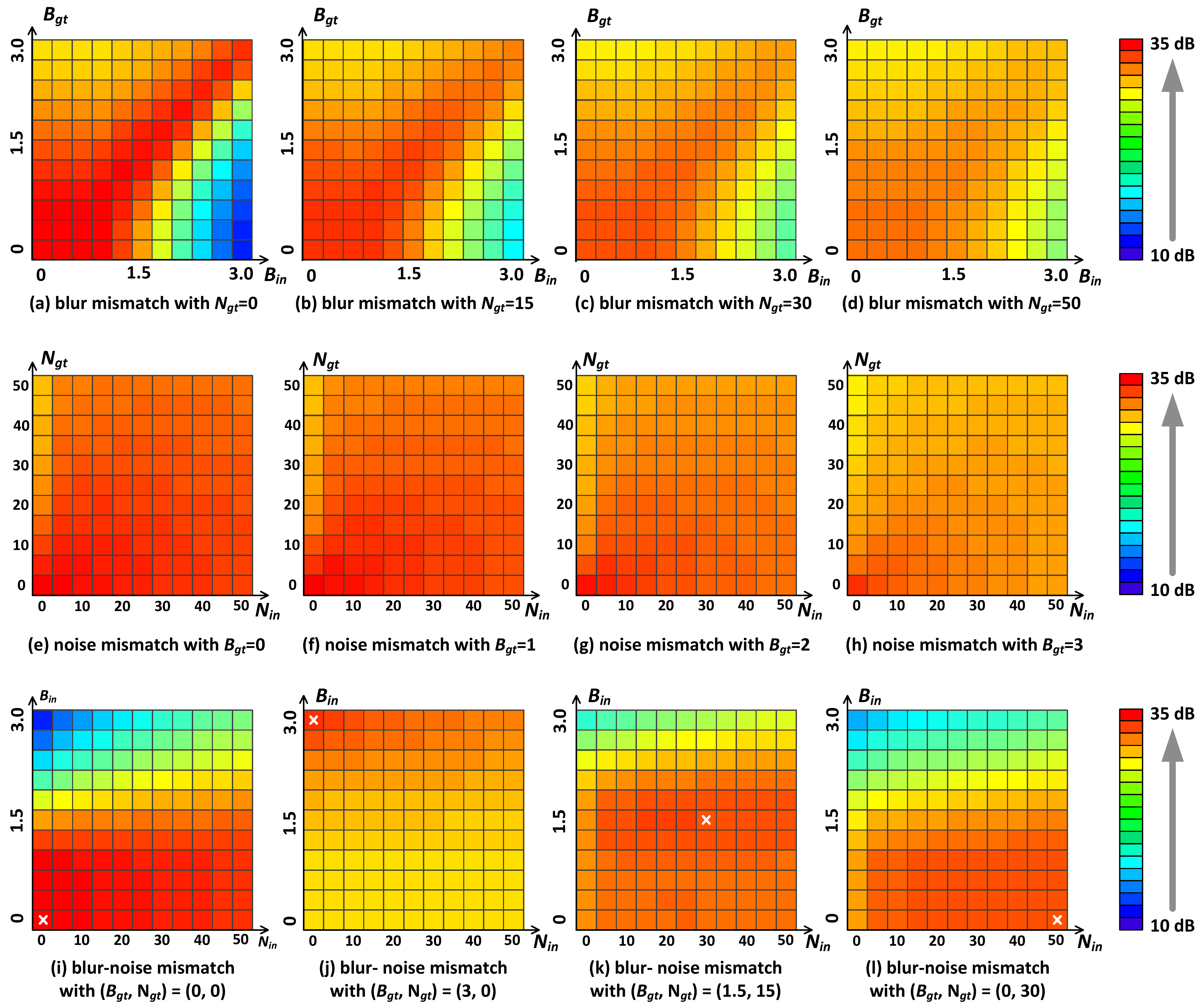}
 \vspace{0.1cm}
 \caption{Visualization of the performance variation of our method with mismatched degradation. (a)-(d) show the PSNR values achieved with mismatched blur kernel widths under different noise levels. (e)-(h) show the PSNR values achieved with mismatched noise level under different blurs. (i)-(l) show the PSNR values achieved with simultaneously mismatched blurs and noise levels under four representative degradation settings (marked by white cross). }\label{fig:mismatch}
 \end{figure*}

\subsubsection{Degradation-Modulating Convolution}
 As the core component of our LF-DMnet, DM-Conv can adapt image features to the given degradation and thus enhances the capability to handle different degradation. As shown in Table~\ref{tab:ablation}, without using DM-Conv, Model 2 suffers from a 1.61 dB decrease in average PSNR as compared to LF-DMnet. This is because, different degradation have different spatial characteristics (as analyzed in Section~\ref{sec:DM-Block}) and cannot be well handled via fixed convolution kernels. In contrast, our DM-Conv dynamically generates convolutional kernels conditioned on the input degradation to recover the degraded image features, and thus achieves higher PSNR values on a wide range of synthetic degradation. Moreover, we visualize the kernels of our DM-Convs (averaged along the channel dimension) with different input blur and noise levels. As shown in Fig.~\ref{fig:visual-kernel}, all the four DM-Convs learn different kernel patterns for different input degradation, and the kernel intensity also varies at different network stages. The above quantitative and visualization results demonstrate the effectiveness of our DM-Conv.

\subsubsection{Angular Information}
 The major difference between our LF-DMnet and non-blind single image SR methods (e.g., SRMD) is the incorporation of the angular information. As shown in Table~\ref{tab:ablation}, when the angular information is not used (i.e., Model 3), the average PSNR value suffers a 2.25 dB drop. This performance gap is also consistent with the gap between SRMD and our method in Table~\ref{tab:efficiency}. This clearly demonstrates that the complementary inter-view correlation is crucial for real-world LF image SR.

\subsubsection{Kernel Prior Embedding}
 It can be observed in Table~\ref{tab:ablation} that Model 4 without kernel prior embedding suffers a 0.22 dB decrease in PSNR as compared to our LF-DMnet, and the PSNR drop is more significant on noise-free scenes. That is because, without KPE, our network has to search for the best degradation kernel to recover the degraded image features. Since we adopt the isotropic Gaussian kernel as the blur kernel for synthetic degradation, KPE can help our network to reduce the searching space and thus facilitates our network to learn more accurate kernel representations.

\subsubsection{Degradation-Modulating Channel Attention}
 As shown in Table~\ref{tab:ablation}, when the degradation-modulating channel attention is removed, Model 5 suffers a 0.20 dB decrease in average PSNR as compared to LF-DMnet. This demonstrates the effectiveness of channel-wise degradation modulation. Since our DM-Conv can only adapt to different degradation in the spatial dimension, DM-CA can be used as a complementary part of DM-Conv to enhance its degradation adaptation capability.
It is also worth noting that our DM-CA only introduces 0.01M increase in model size, 2 ms increase in running time and negligible increase in FLOPs. These results demonstrate the high efficiency of our model design.

\begin{figure*}[t]
 \centering
 \includegraphics[width=18cm]{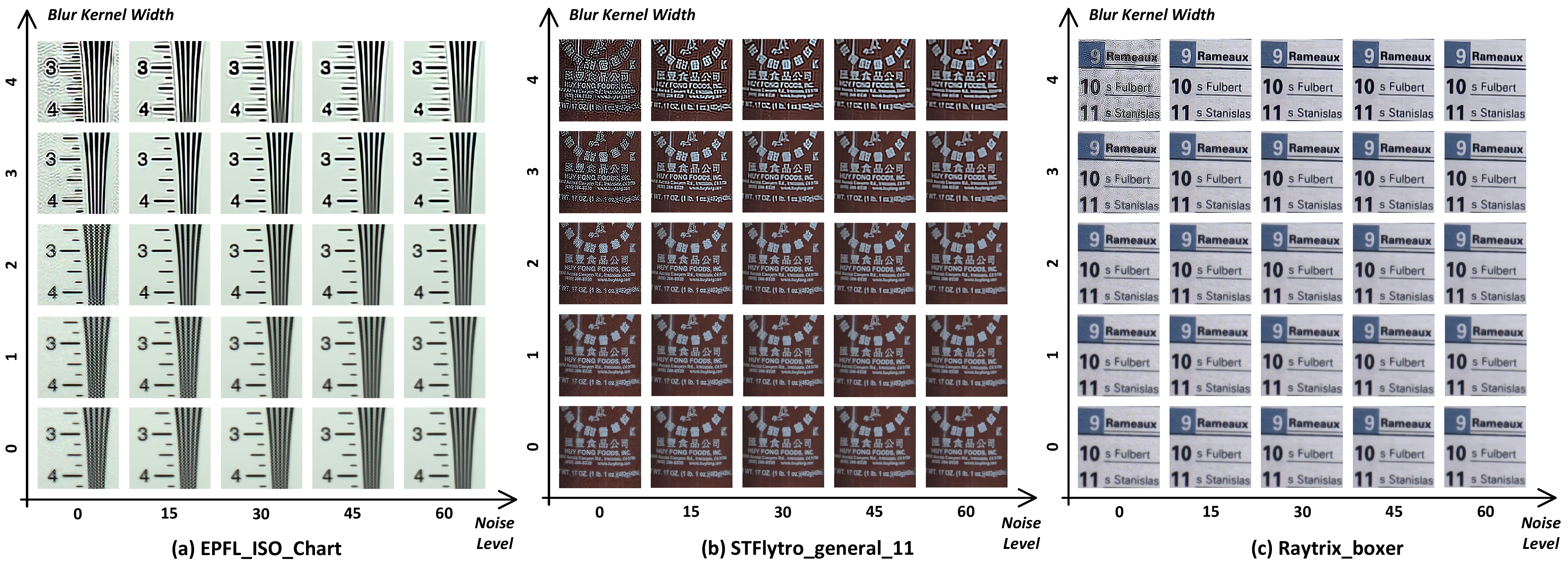}
 \caption{Visual results achieved by our method on real LF images with different blur kernel width and noise levels. Three scenes from the EPFL \cite{EPFL}, STFlytro \cite{STFlytro} and Raytrix \cite{Raytrix} datasets are used as examples for illustration.}\label{fig:vary-blur-noise}
 \end{figure*}

\subsection{Degradation Mismatch Analyses}\label{sec:mismatch}
 In this subsection, we first analyze the performance variation of our method with mismatched input and groundtruth synthetic degradation. Then, we apply our LF-DMnet to real LF images and analyze its SR performance with various input blur kernel widths and noise levels.

 \subsubsection{Synthetic Degradation}
 Since our LF-DMnet is a non-blind SR method, it requires to take the blur kernel and noise level as its input. In the aforementioned experiments with synthetic degradation, we directly use the groundtruth degradation as the input degradation of our network. To investigate the performance of our method when the input degradation mismatches with the groundtruth one, we conduct the following experiments. \textbf{\textit{First}}, we investigate the performance variation of our LF-DMnet with mismatched blur kernel widths by traversing the groundtruth kernel width $B_{gt}$ and the input kernel width $B_{in}$ from 0 to 3 with a step of 0.3. Figures~\ref{fig:mismatch}(a)-\ref{fig:mismatch}(d) visualize the PSNR values achieved by our method (averaged on the validation scenes) under four different noise levels (i.e., $N_{gt}=0, 15, 30, 50$). \textbf{\textit{Second}}, we investigate the performance variation of our LF-DMnet with mismatched noise levels by traversing the groundtruth noise level $N_{gt}$ and the input noise level $N_{in}$ from 0 to 50 with a step of 5. Figures~\ref{fig:mismatch}(e)-\ref{fig:mismatch}(h) visualize the PSNR values achieved by our method (averaged on the validation scenes) under four different blur kernel widths (i.e., $B_{gt}=0, 1, 2, 3$). \textbf{\textit{Third}}, we investigate the performance variation of our LF-DMnet with simultaneously mismatched blur kernel and noise level by traversing the input blur kernel width $B_{in}$ (from 0 to 3 with a step of 0.3) and the input noise level $N_{in}$ (from 0 to 50 with a step of 5). Figures.~\ref{fig:mismatch}(i)-\ref{fig:mismatch}(l) visualize the PSNR values achieved by our method (averaged on the validation scenes) under four representative degradation settings including $\left(B_{gt}, N_{gt}\right)=(0,0), (3,0), (1.5,15), \text{and} (0,30)$.

 From Fig.~\ref{fig:mismatch}, we can draw the following conclusions:
 \begin{itemize}
    \item Best SR performance can be achieved when the input degradation matches the groundtruth one.
    \item The performance variation caused by blur mismatch is more significant than that caused by noise mismatch.
    \item When $B_{in}\neq B_{gt}$,  $B_{in}>B_{gt}$ leads to much more significant performance degradation than $B_{in}<B_{gt}$.
    \item As the noise level increases, the PSNR variation caused by the blur kernel mismatch is reduced.
 \end{itemize}

 \subsubsection{Real-World Degradation}\label{sec:mismatch-real}
 To investigate the influence of the input kernel widths and noise levels to the SR performance under real-world degradation, we directly apply our LF-DMnet to the LFs captured by Lytro and Raytrix cameras, and traverse the input blur kernel width (from 0 to 3 with a step of 1) and the input noise level (from 0 to 60 with a step of 15). Since both the groundtruth HR images and their degradation are unavailable, we evaluate the performance of our method by visually comparing its SR results. Figure~\ref{fig:vary-blur-noise} shows the $4\times$ SR results achieved by our method with varied input degradation, from which we can obtain the following conclusions.
  \begin{itemize}
    \item A large input kernel width can enhance the local contrast and sharpens edges and textures,  but an over-large kernel width introduces ringing artifacts to the result images.
    \item A large input noise level can enhance the local smoothness and helps to alleviate the artifacts, but an over-large input noise level makes the result images blurring.
    \item Our LF-DMnet can achieve better SR performance on Lytro LFs by setting kernel width and noise level to 2 and 30, respectively, and can achieve better SR performance on Raytrix LFs by setting kernel width and noise level to 4 and 60, respectively.
 \end{itemize}
Readers can further refer to our interactive online demo\footnote{\url{https://yingqianwang.github.io/LF-DMnet/}} to view the influence of input degradation to the SR results.

\section{Conclusion and Discussion} \label{sec:Conclusion}
 In this paper, we achieve real-world LF image SR via degradation modulation. We developed an LF degradation model based on the camera imaging process, and proposed an LF-DMnet that can modulate degradation priors into the SR process. Experimental results show that our method can produce visually pleasant and angular consistent SR results on real-world LF images. Through extensive ablation studies and model analyses, we validated the effectiveness of our designs and obtained a series of insightful observations.

 It is worth noting that, although our LF-DMnet achieves significantly improved performance than existing methods on real-world LF image SR, it is sensitive to the input degradation and requires accurate degradation estimation. When the input blur kernel widths and noise levels mismatch with the real ones, our method will produce images with artifacts or over-smoothness. Moreover, due to the non-blind setting in our method, when applying our method to a novel LF camera with unknown degradation, we need to firstly ``measure'' the point spread function and the noise level of this camera, which is user-unfriendly and not practical enough. In the future, we will study the more challenging blind LF image SR problem, and try to design a more practical method for real-world LF image SR. We believe that our LF-DMnet will serve as a fundamental work and can inspire more researchers to focus on real-world LF image SR.
 
 \bibliographystyle{IEEEtran}
 \bibliography{reference}

\begin{IEEEbiography}[{\includegraphics[width=1in,height=1.25in,clip,keepaspectratio]{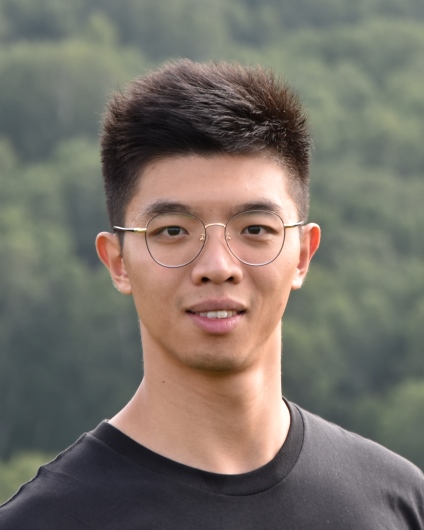}}]{Yingqian Wang} received his B.E. degree in electrical engineering from Shandong University, Jinan, China, in 2016, the Master and the Ph.D. degrees in information and communication engineering from National University of Defense Technology (NUDT), Changsha, China, in 2018 and 2023, respectively. He is currently an assistant professor with the College of Electronic Science and Technology, NUDT. His research interests focus on optical imaging and detection, particularly on light field imaging, image super-resolution and infrared small target detection.
\end{IEEEbiography}

\begin{IEEEbiography}[{\includegraphics[width=1in,height=1.25in,clip,keepaspectratio]{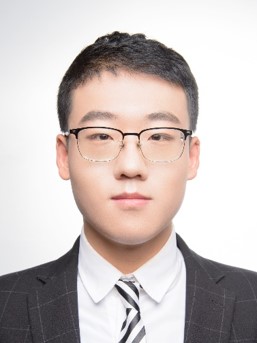}}]{Zhengyu Liang} received the B.E. degree from Xidian University, Xian, China, in 2019, and the M.E. degree in Information and Communication Engineering from National University of Defense Technology (NUDT), Changsha, China, in 2021. He is currently pursuing the Ph.D. degree with the College of Electronic Science and Technology, NUDT. His current research interests mainly focus on low-level vision, particularly on light field image processing and image super-resolution.
\end{IEEEbiography}

\begin{IEEEbiography}[{\includegraphics[width=1in,height=1.25in,clip,keepaspectratio]{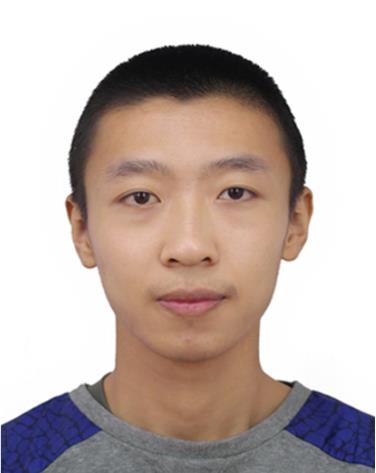}}]{Longguang Wang} received the B.E. degree in Electrical Engineering from Shandong University (SDU), Jinan, China, in 2015, and the Ph.D. degree in Information and Communication Engineering from National University of Defense Technology (NUDT), Changsha, China, in 2022. His current research interests include low-level vision and 3D vision.
\end{IEEEbiography}

\begin{IEEEbiography}[{\includegraphics[width=1in,height=1.25in,clip,keepaspectratio]{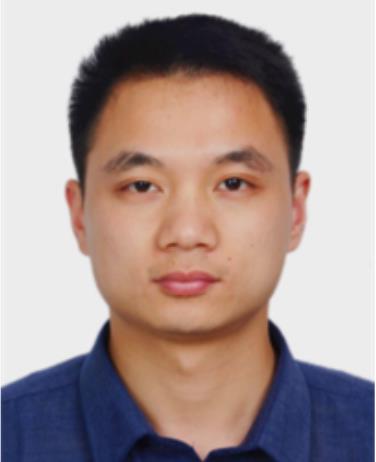}}]{Jungang Yang} received the B.E. and Ph.D. degrees from National University of Defense Technology (NUDT), in 2007 and 2013 respectively. He was a visiting Ph.D. student with the University of Edinburgh, Edinburgh from 2011 to 2012. He is currently an associate professor with the College of Electronic Science, NUDT. His research  interests include computational  imaging, image processing, compressive sensing and sparse representation. Dr. Yang received the New Scholar Award of Chinese Ministry of Education in 2012, the Youth Innovation Award and the Youth Outstanding Talent of NUDT in 2016.
\end{IEEEbiography}

\begin{IEEEbiography}[{\includegraphics[width=1in,height=1.25in,clip,keepaspectratio]{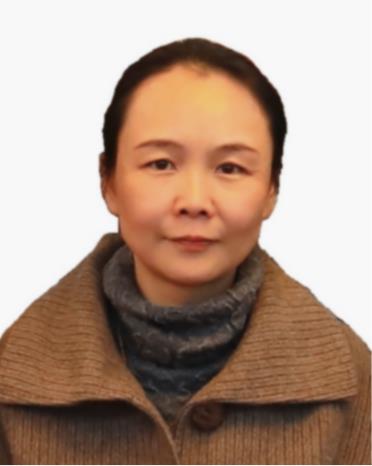}}]{Wei An} received the Ph.D. degree from the National University of Defense Technology (NUDT), Changsha, China, in 1999. She was a Senior Visiting Scholar with the University of Southampton, Southampton, U.K., in 2016. She is currently a Professor with the College of Electronic Science and Technology, NUDT. She has authored or co-authored over 100 journal and conference publications. Her current research interests include signal processing and image processing.
\end{IEEEbiography}

\begin{IEEEbiography}[{\includegraphics[width=1in,height=1.25in,clip,keepaspectratio]{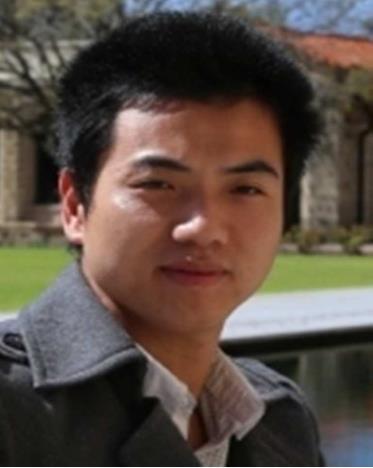}}]{Yulan Guo} is currently an associate professor. He received the B.E. and Ph.D. degrees from National University of Defense Technology (NUDT) in 2008 and 2015, respectively.
He has authored over 100 articles at highly referred journals and conferences. His current research interests focus on 3D vision, particularly on 3D feature learning, 3D modeling, 3D object recognition, and scene understanding. He served as an associate editor for IEEE Transactions on Image Processing, a guest editor for IEEE Transactions on Pattern Analysis and Machine Intelligence, an area chair for CVPR 2021, ICCV 2021, and ACM Multimedia 2021. He organized several tutorials and workshops in prestigious conferences, such as CVPR 2016, CVPR 2019, ICCV 2021, and 3DV 2021. Dr. Guo is a Senior Member of IEEE and ACM.
\end{IEEEbiography}

\end{document}